\documentclass[conference]{IEEEtran}
\IEEEoverridecommandlockouts

\usepackage{url}
\usepackage{amsthm}
\usepackage{afterpage}
\usepackage{cite}
\usepackage[normalem]{ulem}
\useunder{\uline}{\ul}{}
\usepackage{graphicx}
\usepackage{textcomp}
\usepackage{xcolor}
\usepackage{balance} 
\usepackage{bbm}
\usepackage{amsmath,amssymb,amsfonts}
\usepackage{epstopdf}
\usepackage{array}
\usepackage{booktabs}
\usepackage{color}
\usepackage{xcolor}
\usepackage{colortbl}
\usepackage{xspace}
\usepackage{multirow}
\usepackage{pifont}
\usepackage{enumitem}
\usepackage{bbding}
\usepackage{hyperref}
\usepackage{makecell}
\usepackage{microtype}
\usepackage{algorithm}
\usepackage{algorithmic}
\usepackage{fontawesome}
\usepackage{caption,setspace}
\usepackage[utf8]{inputenc}
\newcommand{\ssymbol}[1]{^{\@fnsymbol{#1}}}

\newtheorem{definition}{Definition}
\newtheorem{proposition}{Proposition}

\begin{document}

\title{AdaFGL: A New Paradigm for Federated Node Classification with Topology Heterogeneity}

\makeatletter

\author{
    \IEEEauthorblockN{
    Xunkai Li$^\dagger$, 
    Zhengyu Wu$^\dagger$,
    Wentao Zhang$^{\ddagger \sharp}$, 
    Henan Sun$^\dagger$, 
    Rong-Hua Li$^\dagger$, 
    Guoren Wang$^\dagger$}
    \IEEEauthorblockA{
    $^\dagger$ Beijing Institute of Technology, Beijing, China}
    \IEEEauthorblockA{
    $^\ddagger$ Peking University, $^\sharp$ National Engineering Labratory for Big Data Analytics and Applications, Beijing, China}
    \IEEEauthorblockA{
    cs.xunkai.li@gmail.com, 
    Jeremywzy96@outlook.com,
    wentao.zhang@pku.edu.cn,\\
    magneto0617@foxmail.com,
    lironghuabit@126.com,
    wanggrbit@gmail.com
    }
}

\maketitle

\begin{abstract}
    Recently, Federated Graph Learning (FGL) has attracted significant attention as a distributed framework based on graph neural networks, primarily due to its capability to break data silos.
    Existing FGL studies employ community split on the homophilous global graph by default to simulate federated semi-supervised node classification settings.
    Such a strategy assumes the consistency of topology between the multi-client subgraphs and the global graph, where connected nodes are highly likely to possess similar feature distributions and the same label.
    However, in real-world implementations, the varying perspectives of local data engineering result in various subgraph topologies, posing unique heterogeneity challenges in FGL.
    Unlike the well-known label Non-independent identical distribution (Non-iid) problems in federated learning, FGL heterogeneity essentially reveals the topological divergence among multiple clients, namely homophily or heterophily.
    To simulate and handle this unique challenge, we introduce the concept of structure Non-iid split and then present a new paradigm called \underline{Ada}ptive \underline{F}ederated \underline{G}raph \underline{L}earning (AdaFGL), a decoupled two-step personalized approach.
    To begin with, AdaFGL employs standard multi-client federated collaborative training to acquire the federated knowledge extractor by aggregating uploaded models in the final round at the server.  
    Then, each client conducts personalized training based on the local subgraph and the federated knowledge extractor. 
    Extensive experiments on the 12 graph benchmark datasets validate the superior performance of AdaFGL over state-of-the-art baselines.
    Specifically, in terms of test accuracy, our proposed AdaFGL outperforms baselines by significant margins of 3.24\% and 5.57\% on community split and structure Non-iid split, respectively.

\end{abstract}


\section{Introduction}
\label{sec: introduction}
    Graphs excel at capturing complex relationships between entities and providing a flexible and intuitive representation of them. 
    Such strengths make graphs well-suited for implementation in biomedical~\cite{bang2023app_gnn_bio1, qu2023app_gnn_bio2, gao2023app_gnn_bio3}, recommendation~\cite{xia2023app_gnn_rec1, yang2023app_gnn_rec2, cai2023app_gnn_rec3}, and finance~\cite{balmaseda2023app_gnn_fina1, hyun2023app_gnn_fina2, qiu2023app_gnn_fina3}.
    Recently, Graph Neural Networks (GNNs) have emerged as powerful tools for graph-based data engineering. 
    Through learning from graphs, GNNs achieve great success in various downstream tasks, including node-level~\cite{2016Convolutional_ChebNet,2020h2gcn,velivckovic2017gat}, edge-level~\cite{Zhang18link_prediction1,cai2021link_prediction2,link_prediction3}, and graph-level~\cite{zhang2019graph_classification1,ma2019graph_classification2,yang2022graph_classification3}.

    Despite its high popularity, the practical need for performing graph learning by utilizing the graphs collected from multiple institutions without directly sharing the local graph encourages the emergence of a more distributed framework. 
    The naive approach entails each client independently conducting local training using their own collected graph.
    However, the limited data often results in sub-optimal performance, which brings distributed challenges to existing centralized graph learning methods (e.g., GNNs).
    Motivated by the success of Federated Learning (FL)~\cite{yang2019fl_survey}, Federated Graph Learning (FGL) appears as a feasible solution for graph learning while solving the data silos issues using multi-client collaborative training. 

    During the research phase of FGL, obtaining real-world federated datasets poses challenges due to regulatory limitations~\cite{he2021fedgraphnn,WangFedScope_22_fsg}. 
    Instead, many FGL studies resort to data simulation strategies to generate distributed subgraphs for experiments.
    In particular, two methods, Louvain~\cite{blondel2008louvain} and Metis~\cite{karypis1998metis}, are widely employed for partitioning large graphs into multiple clustered subgraphs to simulate real-world distributed scenarios.
    The recent FGL package~\cite{WangFedScope_22_fsg} refers to them as community split, which represents data simulation strategies based on cluster discovery.
    In this process, an ideal experimental setup is employed, where community split is applied to the homophilous global graph by default, ensuring that the subgraphs owned by multiple clients maintain the same homophilous topology as the global graph. 
    Such homophily assumption~\cite{wu2020gnn_survey1,zhou2022gnn_survey2,bessadok2022gnn_survey3} leads to the exceptional performance of GNNs, which performed as a centralized graph learning technique in each client during the FGL training. 
    In other words, these models consider connected nodes to share similar feature distributions and labels. 
    

    However, the deployment of GNNs in real-world scenarios has revealed the presence of topological heterophily, where connected nodes exhibit contrasting attributes~\cite{ma2021hete_gnn_survey1,luan2022hete_gnn_survey2}.
    The presence of heterophily poses a significant challenge for GNNs, particularly in the FGL with collaborative training, leading to negative effects among multiple clients. 
    These effects stem from the distribution heterogeneity observed in the subgraph homophilous or heterophilous topology, referred to as topology variations.  
    Unlike the Non-iid problem in FL, which primarily pertains to feature or label distributions, the heterogeneity in FGL is inherently tied to the diverse topologies among clients. 

    To address the challenges posed by topology heterogeneity and overcome the limitations of the idealized settings used in the previous FGL studies, we introduce the concept of structure Non-iid split, whose motivation lies in the ubiquity of aforementioned topology heterogeneity challenges in real-world scenarios~\cite{zheng2022hete_gnn_survey3,platonov2023hete_gnn_survey4}. 
    This data simulation strategy aims for providing a new benchmark for future FGL studies and bridging the gap between the research and real-world implementations.
    For example, the research team-based citation network exhibits relationships representing intersectional fields (e.g., AI4Science)~\cite{pfeifer2023fedapp_gnn_bio1,wufederated2023fedapp_gnn_bio2}.
    In detecting fraudulent actions within online transaction networks, fraudsters are more likely to build connections with customers~\cite{pan2022fedapp_gnn_fina1,cao2022fedapp_gnn_fina2}. 
    The patient networks exhibit varying connection patterns due to regional differences and numerous disease genres~\cite{pfeifer2022fedapp_gnn_dis1,ahmed2022fedapp_gnn_dis2}.

\begin{figure}[t]
  \includegraphics[width=\linewidth]{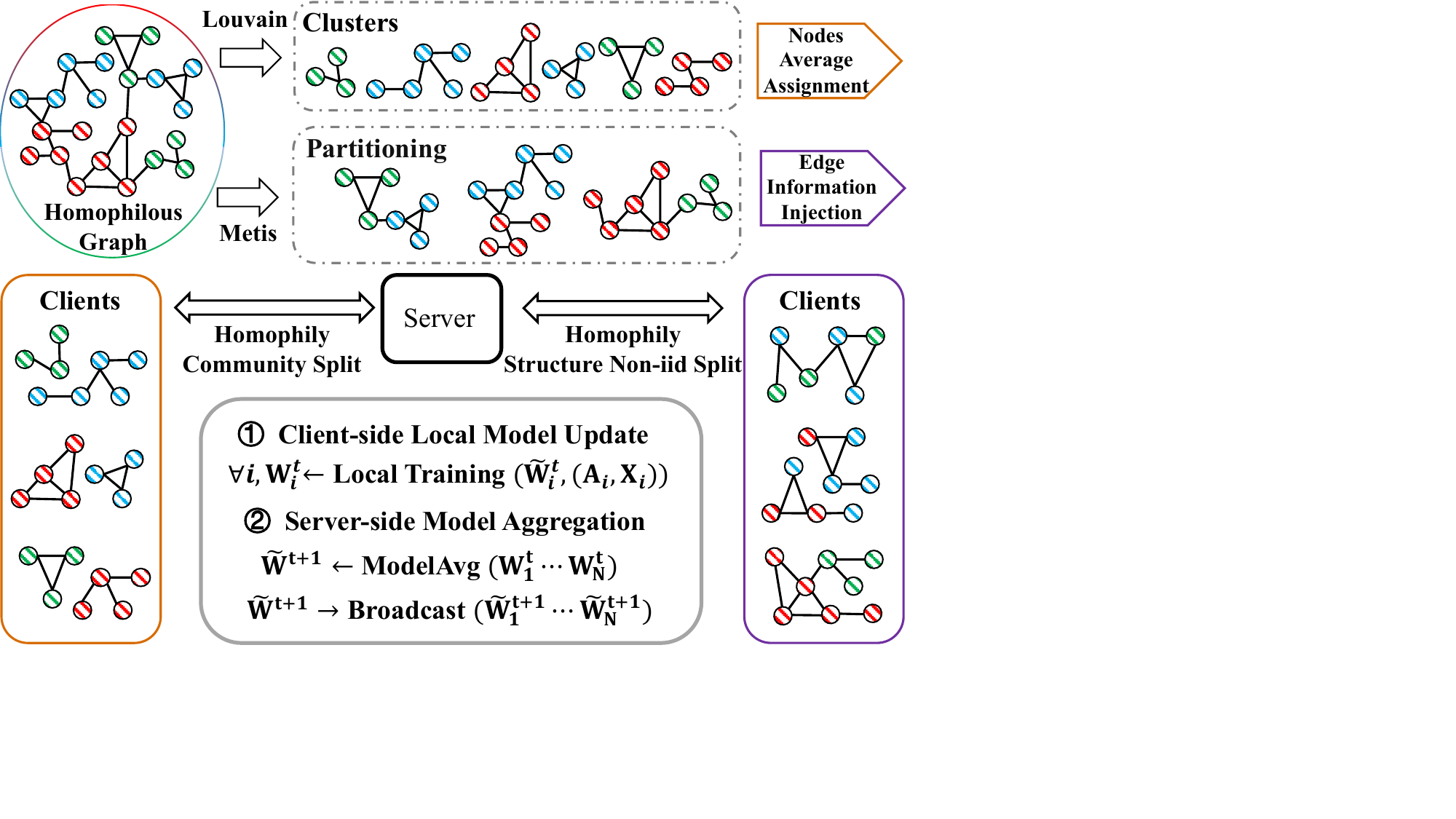}
  \captionsetup{font={small,stretch=1}}
  \caption{
  Overview of standard FGL pipeline with two data simulation strategies in the same homophilous global graph. 
  The different colors of the nodes represent the different labels.}
  \vspace{-0.6cm}
  \label{fig:motivation_homo_hete}
\end{figure}

\begin{figure*}[t]
  \includegraphics[width=\textwidth]{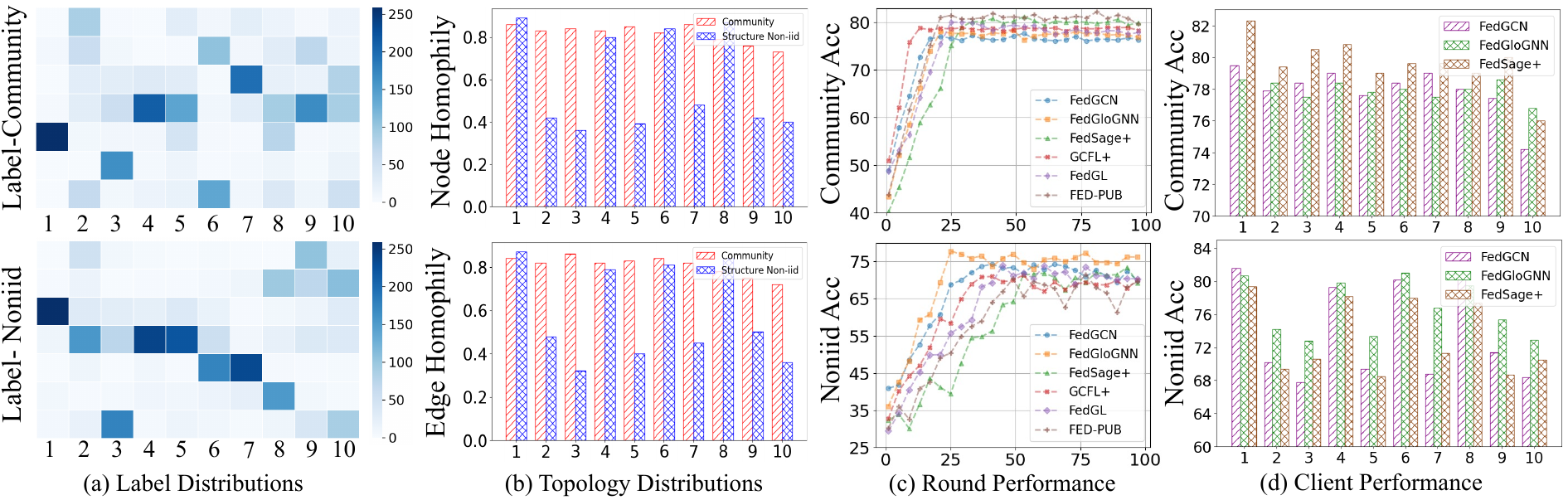}
    \captionsetup{font={small,stretch=1}}
  \caption{
   The empirical analysis is based on the Cora with 10 client community split and structure Non-iid split. 
   (a) the color from white to blue represents the number of nodes held by different clients in each class. 
   (b) quantifying the topology of multi-client subgraphs from both node and edge perspectives, where higher values indicate stronger structural homophily. 
   Please refer to Sec.~\ref{sec: Preliminaries} for detailed computation of the metrics.
   (c) The x-axis of the line plot represents the federated training round. 
   (d) The x-axis of the bar plot represents the client ID.}
  \vspace{-0.25cm}
  \label{fig:motivation_exp}
\end{figure*}

    To further illustrate the differences between the two data simulation strategies mentioned above, we provide an example shown in Fig.~\ref{fig:motivation_homo_hete}. 
    In the community split, Louvain identifies homophilous communities and then allocates them to clients using the  node average assignment principle.
    In the structure Non-iid split, Metis identifies subgraphs corresponding to the number of clients.
    To ensure topology variations, we then apply edge information injection to each client, introducing a binary process to enhance either homophily or heterophily.
    To further discuss the impact of the two above stimulation strategies, we present a comprehensive empirical analysis in Fig.~\ref{fig:motivation_exp}, aiming to address the following questions: 
    \textbf{Q1}: Compared with the well-known Non-iid problem in FL, what are the challenges posed by heterogeneity in FGL that most contribute to the negative impact on performance?
    \textbf{Q2}: How do existing methods perform when confronted with the challenges of heterogeneity in FGL?

    To answer \textbf{Q1}, we first need to clarify the definition of the Non-iid problem in the traditional federated learning settings. 
    It refers to varying feature or/and label distribution held by each client based on its collected data sample, rejecting the premise that all data comes from the same distribution. 
    These differences may arise from factors such as various geographical locations and device differences.
    From the data engineering perspective, if there is a correlation between the data features and labels, the Non-iid problem often manifests as the existence of significant differences in the label distributions across multiple clients.

    In Fig.~\ref{fig:motivation_exp}(a), we demonstrate the label distributions of multiple clients after applying community split and structure Non-iid split to Cora, which serves as an example of the label Non-iid.
    Due to the homophily exhibited by Cora, connected nodes identified by Louvain and Metis algorithms within the same client are likely to possess the same label. 
    As a result, the densely connected modules contain only a few categories, as depicted in Fig.~\ref{fig:motivation_homo_hete}.
    In contrast, when the global graph topology displays heterophily, the label distribution will become more uniform under the same community split. 
    Thus, relying solely on feature or label distributions for analysis becomes unreliable, as they vary correspondingly to the topology divergence of the global graph.
    To further investigate, Fig.~\ref{fig:motivation_exp}(b) provides quantitative evidence of the heterogeneity challenges in FGL by examining topology distribution among clients in both scenarios.
    Compared to the consistent subgraph topologies in community split, structure Non-iid split generates diverse topology distributions.
    Moreover, compared to the label Non-iid problem, which becomes vague as the global graph exhibits homophily or heterophily, the topological distribution indicated under the setting of structure Non-iid split consistently reveals the existence of FGL heterogeneity. 
    Therefore, we propose that FGL heterogeneity is fundamentally influenced by topology variations, and the topology variations generated through structure Non-iid split are more suitable for analysis.

    To answer \textbf{Q2}, we evaluate the predictive performance of the recently-proposed FGL approaches, including FedSage+~\cite{zhang2021fedsage}, GCFL+~\cite{xie2021gcfl}, FedGL~\cite{chen2021fedgl}, and FED-PUB~\cite{baek2022fedpub}. 
    Additionally, we include two federated implementations of representative GNNs: FedGCN~\cite{kipf2016gcn}, which serves as a simple yet effective baseline model for homophilous graphs, and FedGloGNN~\cite{2022glognn}, a state-of-the-art method designed to tackle the heterophilous topology challenges in the central graph learning. 
    Based on the performance curves presented in Fig.~\ref{fig:motivation_exp}(c),  all methods exhibit relatively stable convergence in the community split scenario. 
    In contrast, the topology heterogeneity caused by structure Non-iid split makes all baselines difficult to converge and results in unstable and sub-optimal predictive performance. 
    Specifically, while both FedSage+ and FED-PUB demonstrate competitive performance in the community split scenario, the presence of topology heterogeneity poses a disaster to them.
    Meanwhile, due to the advantage of GloGNN in handling heterophilous topology, FedGloGNN performs best in the structure Non-iid split scenario but fails to achieve competitive results in the community split compared to FedGCN.
    This indicates that heterophilous GNNs in FGL still have room for improvement.

\begin{figure}[t]
  \includegraphics[width=\linewidth]{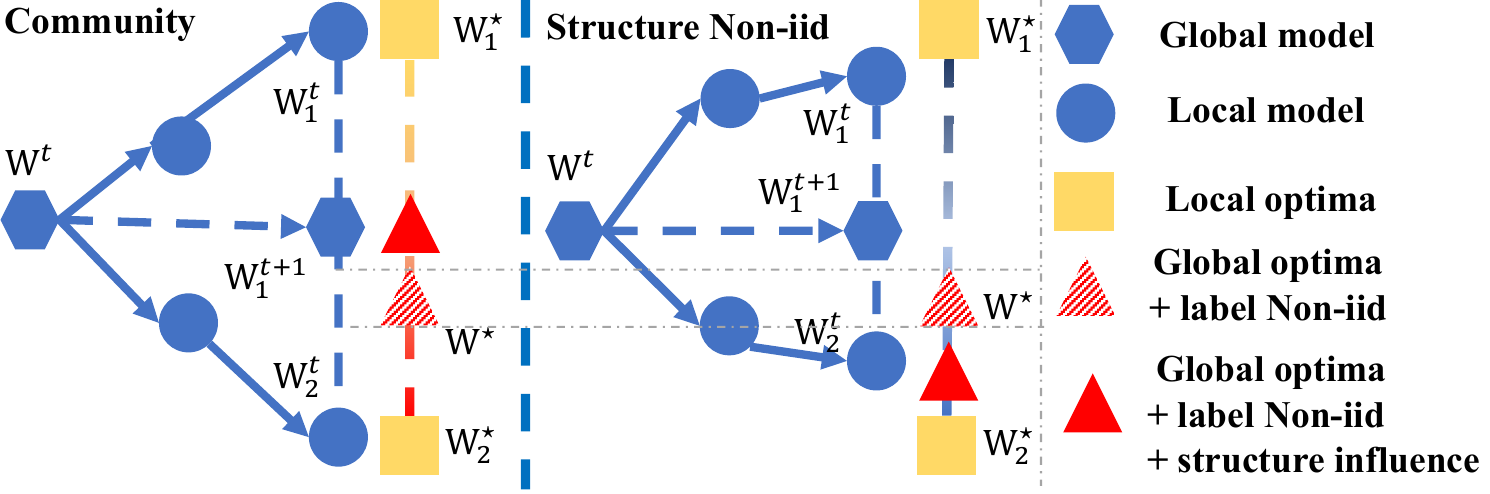}
    \captionsetup{font={small,stretch=1}}
  \caption{
   A toy example illustrating the impact of topology heterogeneity on FGL with the homophilous global graph.}
  \vspace{-0.6cm}
  \label{fig:motivation_optima}
\end{figure}

    To further illustrate this issue, we present the performance of each client in Fig.~\ref{fig:motivation_exp}(d). 
    We observe that, in the community split, most baselines perform better on subgraphs with single-category distributions (Fig.~\ref{fig:motivation_exp}(a), darker colors) indicating strong homophily (Fig.~\ref{fig:motivation_exp}(b), higher Node/Edge Homophily). 
    This contradicts our intuition regarding the label Non-iid problem in FL, which leads to drift between local and global models, resulting in the poor predictive performance of the global model on local data. 
    Interestingly, we observe that in the structure Non-iid split, even under a single-category distribution, baselines fail to achieve satisfactory performance when the subgraph topology exhibits topology variations.
    To explain it, we propose Proposition.~\ref{proposition:fgl_homo_hete}, which aligns with our results.
    Meanwhile, we provide a simple yet direct illustration in Fig.~\ref{fig:motivation_optima}.

    \begin{proposition}
    \label{proposition:fgl_homo_hete}
    Among multiple clients of FGL, topological homophily attracts both the global model and optima, while topological heterophily diverges the global model and optima.
    \end{proposition}

    To tackle this topology heterogeneity, we propose \underline{\textbf{Ada}}ptive \underline{\textbf{F}}ederated \underline{\textbf{G}}raph \underline{\textbf{L}}earning (AdaFGL), a two-step paradigm.
    \textbf{Step1}: 
    AdaFGL employs standard federated training to obtain the federated knowledge extractor, which is derived from aggregating uploaded local models in the final server round.
    Subsequently, each client leverages the client-shared federated knowledge extractor to optimize their local topology.
    \textbf{Step2}: 
    Each client achieves local personalized training based on the optimized local topology.
    In summary, our main contributions are as follows: 
    (1) \textbf{New Perspective}. 
    To the best of our knowledge, this paper is the first to investigate FGL heterogeneity, which we propose to be fundamentally related to topology variations among multiple clients.
    To simulate it, we introduce the structure Non-iid split, which provides a new benchmark for FGL. 
    (2) \textbf{New Paradigm}.
    We propose AdaFGL, a user-friendly and flexible paradigm. 
    It effectively addresses the local optimization dilemma resulting from topology variations among clients, while also minimizing communication costs and the risk of privacy breaches through personalized technologies. 
    (3) \textbf{State-of-the-art Performance}. 
    Experimental results demonstrate that AdaFGL achieves SOTA performance
    on 12 datasets with two data simulation methods.
    Specifically, AdaFGL achieves 2.92\% and 6.28\% performance gains in the homophilous and heterophilous datasets, respectively.

\section{Preliminaries}
\label{sec: Preliminaries}


\subsection{{Notation and Problem Formalization}}
\label{sec: Notation and Problem Formalization}
    Consider a graph $G = (\mathcal{V}, \mathcal{E})$ with $|\mathcal{V}|=n$ nodes and $|\mathcal{E}|=m$ edges, the adjacency matrix (including self-loops) is $\hat{\mathbf{A}}\in\mathbb{R}^{n\times n}$, the feature matrix is $\mathbf{X} = \{x_1,\dots,x_n\}$ in which $x_v\in\mathbb{R}^{f}$ represents the feature vector of node $v$, and $f$ represents the dimension of the node attributes.
    Besides, $\mathbf{Y} = \{y_1,\dots,y_n\}$ is the label matrix, where $y_v\in\mathbb{R}^{|\mathcal{Y}|}$ is a one-hot vector and $|\mathcal{Y}|$ represents the number of the classes.
    The semi-supervised node classification task is based on the topology of labeled set $\mathcal{V}_L$ and unlabeled set $\mathcal{V}_U$, and the nodes in $\mathcal{V}_U$ are predicted based on the model supervised by $\mathcal{V}_L$.
    {In the context of FGL, there is a trusted server and $n$ clients. 
    During the collaborative training process, each client performs semi-supervised node classification based on the locally private subgraph $G_i$.
    Notably, there is a lack of federated graph benchmark datasets due to privacy regulations.
    Therefore, existing FGL methods~\cite{chen2021fedgl,zhang2021fedsage,baek2022fedpub} simulate federated setting through community split on the global graph to maintain consistent topology in multi-client subgraphs (homophily or heterophily).
    Considering the intricate real-world FGL application scenarios and structural noise introduced by various multi-client data collection methods and qualities, we propose the structure Non-iid split for a more generalized FGL setting. 
    Its formal definition is as follows:
    \begin{definition}
    \label{definition:structure_non-iid_split}
    (Structure Non-iid split) 
    Consider a graph $G(V, E)$. 
    We first apply $n$-client Metis on $G$ to obtain federated subgraphs $G_1(V_1, E_1), \dots, G_n(V_n, E_n)$, which have topological consistency with $G$. 
    Then, we perform binary selection on each subgraph to inject homophilous or heterophilous edges $\tilde{E}_i$.
    Eventually, we obtain $\tilde{G}_1(V_1, \tilde{E}_1), \dots, \tilde{G}_n(V_n, \tilde{E}_n)$ for FGL.
    \end{definition} 
    Based on this, we aim to propose an FGL approach that can achieve competitive performance in both the community split and structure Non-iid split.
    Moreover, we intend to utilize the structure Non-iid split to evaluate existing methods and establish new benchmark tests for future FGL research.
    }

\subsection{Prior GNNs in Central Graph Learning}
\label{sec: prior gnns in central graph learning}
    Motivated by the spectral graph theory and deep learning, the graph convolution is initially proposed in~\cite{2013firstgnn}.
    Building upon this, Graph Convolutional Network (GCN)~\cite{kipf2016gcn} simplifies the topology-based convolution operator by the first-order approximation of Chebyshev polynomials, which operates on the widely adopted homophily assumption in the central graph learning.
    Specifically, GCN iteratively propagates node attribute information to adjacent nodes during learning.
    The forward propagation process of the $l$-th layer in GCN is formulated as
    \begin{equation}
        \label{eq:gcn}
        \mathbf{X}^{(l)}\! =\! \sigma(\tilde{\mathbf{A}}\mathbf{X}^{(l-1)}\mathbf{W}^{(l)}),\;\tilde{\mathbf{A}} \!=\! \hat{\mathbf{D}}^{r-1}\hat{\mathbf{A}}\hat{\mathbf{D}}^{-r},\;r\in[0,1],
    \end{equation}
    where $\hat{\mathbf{D}}$ is the degree matrix of $\hat{\mathbf{A}}$, $r$ is the convolution kernel coefficient, $\mathbf{W}$ is the trainable weights, and $\sigma(\cdot)$ is the non-linear activation function.
    By setting $r$, we can get the random walk $\hat{\mathbf{A}}\hat{\mathbf{D}}^{-1}$~\cite{xu2018jknet} and  the reverse transition $\hat{\mathbf{D}}^{-1}\hat{\mathbf{A}}$~\cite{zeng2019graphsaint} variant.
    In GCN, by setting $r=1/2$, we can acquire $\hat{\mathbf{D}}^{-1/2}\hat{\mathbf{A}}\hat{\mathbf{D}}^{-1/2}$.
    By utilizing it, some recent studies~\cite{hamilton2017graphsage,2019appnp,wang2020gcnlpa,chen2020gcnii} optimize the model architectures to improve performance. 
    There are also methods that focus on the propagation process to encode deep structural information.
    For example, SGC~\cite{wu2019sgc} utilizes a linear model operating on $k$-layer propagated features: $\mathbf{X}^{(k)}=\tilde{\mathbf{A}}^k \mathbf{X}^{(0)}$.
    GAMLP~\cite{gamlp} achieves message aggregation based on the attention mechanisms: $\widetilde{\mathbf{X}}_i^{(l)}=\mathbf{X}_i^{(l)} \| \sum_{k=0}^{l-1} w_i(k) \mathbf{X}_i^{(k)}$, where attention weight $w_i(k)$ has multiple calculation versions.

    Despite effectiveness, a recent survey on GNNs~\cite{ma2021hete_gnn_survey1} reveals their limitations under real-world application scenarios since the homophily assumption. 
    In detail, there are two-folded quantitative metrics for measuring the topological homophily: node homophily~\cite{pei2020geomgcn} and edge homophily~\cite{2020h2gcn}
        \begin{equation}
        \label{eq: homo_metric}
          \begin{aligned}
            &\mathcal{H}_{node}=\frac{1}{n} \sum_{v \in \mathcal{V}} \frac{\left|\left\{u \in \mathcal{N}_v: y_v=y_u\right\}\right|}{|\mathcal{N}_v|},\\
            &\mathcal{H}_{edge}=\sum_{e\in \mathcal{E}}\frac{\left|\left\{(v, u) \in {e}: y_v=y_u\right\}\right|}{m},
          \end{aligned}
        \end{equation}
    where $\mathcal{N}_v$ represents the one-hop neighbors of node $v$.
    This limitation arises in message aggregation, as it depends on homophily for feature augmentation.
    However, the presence of heterophily misleads this process.
    Recent approaches aim to capture heterophily by incorporating higher-order neighbor discovery or improved message combination. 
    GPR-GNN~\cite{chien2021gprgnn} controls the contribution of propagated features in each step by learnable generalized PageRank $\mathbf{Z}=\sum_{k=0}^K \gamma_k \mathbf{X}^{(k)}$.
    GloGNN~\cite{2022glognn} utilizes transformation coefficient matrix $\mathbf{T}$ to generates a node’s global embedding $(1-\gamma)\mathbf{T}^{(l)}\mathbf{X}^{(l)}+\gamma \mathbf{X}^{(l)}$.
    LW-GCN~\cite{dai2022lwgcn} introduces a label-wise message-passing weight $\mathbf{\omega}_{v,k}$ for node $v$ and class $k$ to mitigate the negative impacts arising from aggregating dissimilar neighbors $\mathbf{X}^{(l)} = \sigma(\mathbf{W}^{(l)}\mathrm{CONCAT}(\mathbf{X}^{(l-1)},\mathbf{\omega}_{v,1},\dots,\mathbf{\omega}_{v,k})$).
    The core of the above methods is to break the limitation of first-order aggregation and acquire more comprehensive messages from a global perspective. 
    Inspired by this, several recent approaches~\cite{2020h2gcn,pei2020geomgcn,2021linkx,yan2021ggcn,du2022gbkgnn,song2023ordergnn} further improve model performance through well-designed model architectures and aggregation methods.

\subsection{Federated Graph Learning}
\label{sec: fgl}
    Motivated by the success of the FL in computer vision and natural language processing~\cite{yang2019fl_survey} with the demands for distributed graph learning in the real world, FGL has received growing attention from researchers. 
    To achieve collaborative training for multi-clients, FedAvg~\cite{mcmahan2017fedavg} is proposed.
    Here we illustrate the GNNs combined with FedAvg.
    Its generic form with $i$-th client and learning rate $\eta$ is defined as 
    \begin{equation}
        \small
        \label{eq:localtrain}
        \begin{aligned}
        \mathbf{W}_i^{t}&=\widetilde{\mathbf{W}}_i^{t}-\eta\nabla f\left(\widetilde{\mathbf{W}}_i^t,(\mathbf{A}_i,\mathbf{X}_i,\mathbf{Y}_i)\right)\\
        &=\widetilde{\mathbf{W}}_i^{t}+\eta\sum_{i \in \mathcal{V}_L} \sum_j \mathbf{Y}_{i j} \log \left(\operatorname{Softmax}(\hat{\mathbf{Y}})_{i j}\right),
        \end{aligned}
    \end{equation}
    where $\eta$ denotes the learning rate, $\nabla f(\cdot)$ denotes the gradients.
    $\mathbf{W}_i^t$ and $\widetilde{\mathbf{W}}_i^t$ represent the $i$-th local model in round $t$ and the aggregated global model received from the server.
    The model aggregation weights used in the FedAvg are proportional to the client's data size.
    Based on this, the server executes model aggregation in each communication round. 
    Then, the aggregated global model is broadcast to the participating clients for the next round. 
    The above process can be defined as
    \begin{equation}
    \label{eq:fedavg}
        \begin{aligned}
        \forall i,\;\mathbf{W}^{t}_{i} \leftarrow \widetilde{\mathbf{W}}^{t}_{i} - \eta\nabla f,\;\widetilde{\mathbf{W}}^{t+1} \leftarrow \sum_{i=1}^N\frac{{n}_i}{{n}}\mathbf{W}^{t}_i,
        \end{aligned}
    \end{equation}
    where ${n}_i$ and ${n}$ represent the $i$-th client and total data size.

\begin{figure*}[t]
  \includegraphics[width=\textwidth]{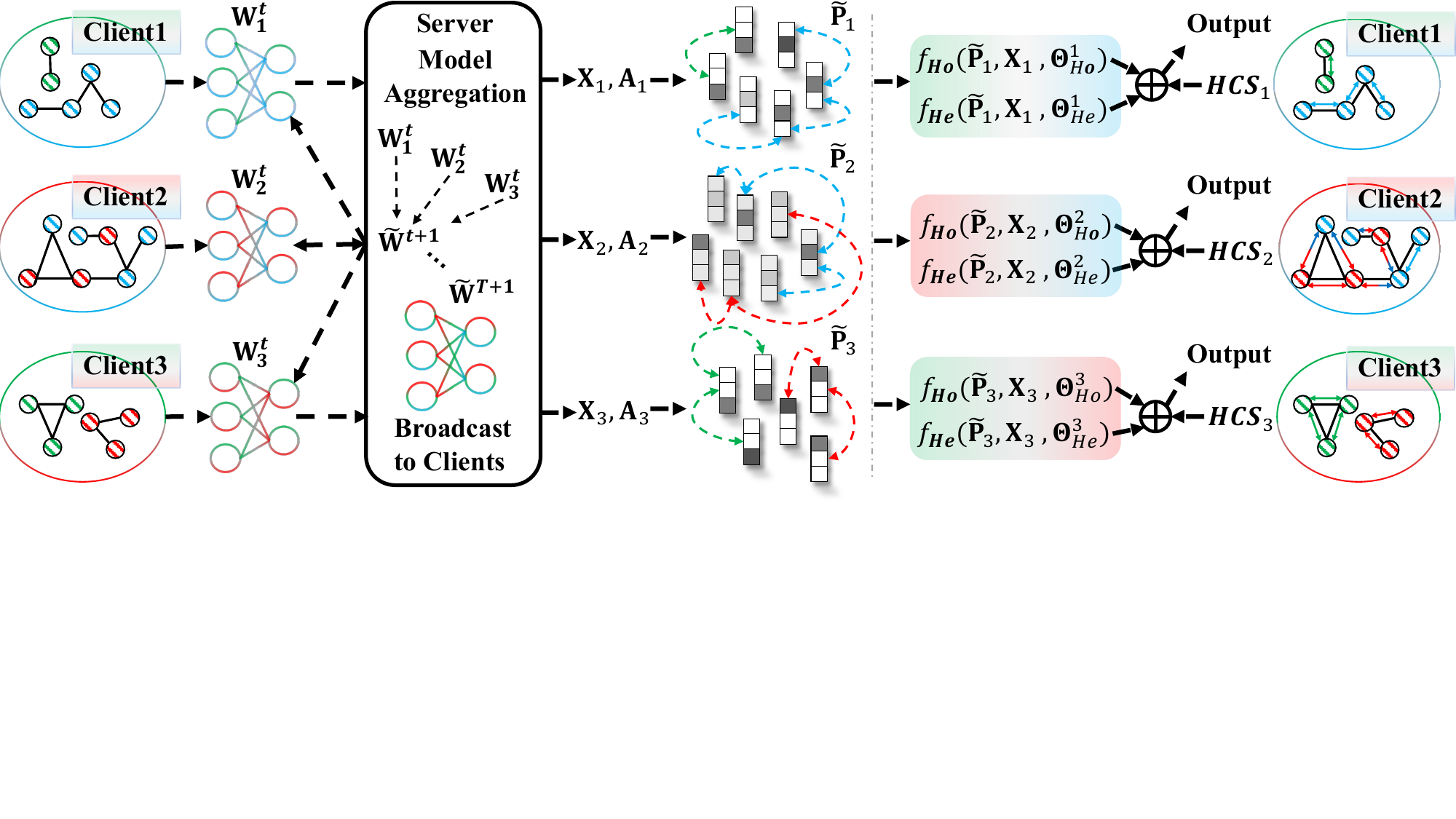}
  \captionsetup{font={small,stretch=1}}
  \caption{
  Overview of the AdaFGL.
  Step 1: standard federated collaborative training, where the federated knowledge extractor parameterized by $\widetilde{\mathbf{W}}^{T+1}$ is obtained by aggregating the last training round's uploaded models at the server and broadcasted to each client.
  Then, each client employs the federated knowledge extractor to optimize the local topology and obtain the optimized probability propagation matrix $\widetilde{\mathbf{P}}$.
  Step 2: each client executes the homophilous/heterophilous propagation module $f_{Ho}$ and $f_{He}$. 
  Subsequently, each client adaptively combines the results of the above two modules using the HCS to obtain the final predictions.
  More training and design details can be found in Sec.~\ref{sec: adaptive federated graph learning}.}
  \label{fig: framework}
  \vspace{-0.4cm}
\end{figure*}

    Recent FGL studies aim to improve model performance under community split via well-designed client-side model architectures, optimization of server-side model aggregation, and establishment of cross-client interactions.
    In FedGL~\cite{chen2021fedgl}, each client uploads the local predictions and embeddings to the server for generating global supervised information, then the server shares this information with each client for local training.
    FedGNN~\cite{wu2021fedgnn} is proposed for federated graph recommendation, it enhances the subgraph representing the interaction history of the local user through random sampling of interacted items and cross-client interactions. 
    FedSage+~\cite{zhang2021fedsage} trains NeighGen to achieve local subgraph augmentation by constructing the multi-client interactions loss function.
    GCFL+~\cite{xie2021gcfl} discovers potential clusters among clients by utilizing the model gradients uploaded by clients and then performs the personalized model aggregation for each discovered cluster. 
    FED-PUB~\cite{baek2022fedpub} focuses on the joint improvement of the interrelated local GNNs. 
    It utilizes the client-side model weight similarity to perform weighted server-side model aggregation. 
    Further, it learns a personalized sparse mask at each client to select and update only the subgraph-relevant subset of the aggregated parameters.

\section{Adaptive Federated Graph Learning}
\label{sec: adaptive federated graph learning}
    In this section, we introduce AdaFGL for federated semi-supervised node classification.
    As a flexible two-step paradigm, AdaFGL is user-friendly, and we provide one simple implementation of its infinite evolution.
    Specifically, in Sec.~\ref{sec: architecture overview}, we present an overview of the AdaFGL pipeline depicted in Fig.~\ref{fig: framework}, followed by the design intuition behind its two-step process.
    Step 1 involves standard federated collaborative training between the server and multiple clients, and each client optimizes their local topology based on the federated knowledge extractor obtained through the server-side aggregation. 
    The details of obtaining the federated knowledge extractor and local topology correction process are provided in Sec.~\ref{sec: federated knowledge extractor}.  
    Step 2 includes personalized propagation on each client using homophily and heterophily propagation modules. 
    Moreover, we also incorporate adaptive combination techniques based on the local topology and avoid manual tuning. 
    Implementation details are provided in Sec.~\ref{sec: local personalized propagation} and Sec.~\ref{sec: homophily confidence score}.

\subsection{Architecture Overview} 
\label{sec: architecture overview}

    According to the empirical analysis in Sec.~\ref{sec: introduction}, we observe the challenges in real-world scenarios imposed by topology heterogeneity and the limitation of existing FGL approaches in addressing such challenges. 
    Therefore, we aim to propose a unified framework that can effectively handle topology heterogeneity while maintaining competitive performance under the community split.
    The following content elucidates the intuition of our proposed two-step decoupling framework.

    \underline{\emph{Step 1: Federated Knowledge Extractor}}: 
    In order to address the FGL optimization dilemma posed by topology variations and the significant impact on predictions shown in Fig.~\ref{fig:motivation_exp} and Fig.~\ref{fig:motivation_optima}, we propose a personalized approach for each client within AdaFGL.
    {This personalized approach considers federated training as a supplementary step and aims to provide tailored solutions for each client.
    Similar personalized strategies have been shown theoretically to be effective in previous related FL studies~\cite{fallah2020personalized1,collins2021personalized2,zhang2021personalized3,acar2021personalized4,tan2022personalized5}.
    In other words, the personalized local representation based on the aggregated global model can approximate the optimum.
    In AdaFGL, we first conduct standard federated training to obtain a federated knowledge extractor on the server, which is then broadcast to each client.
    Subsequently, each client utilizes it to derive an optimized probability propagation matrix.
    Notably, during the standard federated training, the clients and the server only exchange gradient information, minimizing the risk of privacy breaches.}

    \underline{\emph{Step 2: Adaptive Personalized Propagation}}: 
    As Fig.~\ref{fig:motivation_homo_hete} and Fig.~\ref{fig:motivation_exp} demonstrate using a fixed propagation rule across multi-client subgraphs hampers the performance when clients exhibit diverse topologies (i.e., topology heterogeneity).
    To overcome these challenges, we introduce homophilous and heterophilous propagation modules for each client, enabling the generation of comprehensive embeddings that consider the holistic topology (i.e., homophily and heterophily).
    Furthermore, we achieve the adaptive combination of these embeddings by leveraging quantified information learned from local topology, which avoids the need for manual hyperparameter tuning. 

    As mentioned above, as a flexible and novel paradigm, AdaFGL can benefit from advancements in FL optimization and GNNs to obtain a more powerful federated knowledge extractor (Step 1). 
    Meanwhile, researchers have the flexibility to replace components within the propagation modules with alternative designs (Step 2), enabling customization based on specific requirements.
    Furthermore, AdaFGL prioritizes maximizing the computational capacity of the local system, while simultaneously minimizing communication costs and privacy risks.
    To provide a comprehensive understanding of AdaFGL, we present detailed illustrations in Alg.~\ref{alg: adafgl-step1} and Alg.~\ref{alg: adafgl-step2}.

\subsection{Federated Knowledge Extractor}
\label{sec: federated knowledge extractor}
    Essentially, the federated knowledge extractor is derived from the global model obtained through the final aggregation round on the server, which can be viewed as knowledge extracting. 
    In our implementation, we utilize a simple GNN such as GCN and employ FedAvg as the aggregation strategy. 
    While more advanced local GNNs and optimization strategies may yield better performance, the focus of this paper is to propose a general framework for federated node classification rather than achieving performance rankings.

    Based on the aforementioned federated collaborative training in the Sec.~\ref{sec: fgl}, the federated knowledge extractor $\widetilde{\mathbf{W}}^{T+1}$ is obtained after $T$ rounds through Eq.~(\ref{eq:localtrain}) and Eq.~(\ref{eq:fedavg}), and it is shared among multiple clients. 
    By leveraging the local subgraph and the federated knowledge extractor, the $i$-th client enhances the performance of local propagation by incorporating the optimized topology.
    The above process is defined as
    \begin{equation}
    \label{eq:correct_local_topology}
      \begin{aligned}
      \mathbf{P}_{i} = \alpha\mathbf{A}_i+\left(1-\alpha\right)\hat{\mathbf{P}}_i\hat{\mathbf{P}}_i^\mathrm{T},\hat{\mathbf{P}}_i=f\left(\mathbf{X}_i,\mathbf{A}_i,\widetilde{\mathbf{W}}^{T+1}\right),
    \end{aligned}
    \end{equation}
    where $\mathbf{P}_{i}$ represents the corrected probability propagation matrix.
    Since the dense nature and the lack of standardization of the original probability propagation matrix easily lead to high bias, we improve it by scaling the aggregated messages
    \begin{equation}
    \label{eq:scaleA}
      \begin{aligned}
        \widetilde{\mathbf{P}}_{i} = \mathbf{P}_{i} / \mathbf{d}\mathbf{d}^\mathrm{T} - \mathrm{diag}\left( \mathbf{P}_{i}\right).
      \end{aligned}
    \end{equation}
    Formally, let $p_{ij}\in \mathbf{P}$ correspond to the $i$-th row and $j$-th col of $\mathbf{P}$, the scaling operator is $d_{ij} = \mathrm{dis}(p_{ii}, p_{ij})$ for $j \neq i$, where $\mathrm{dis}(\cdot)$ is a distance function or any function positively relative with the difference.
    In our implementation, we utilize identity distance to perform degree normalization.

\begin{algorithm}[t]
\caption{AdaFGL-Federated Knowledge Extractor} 
\label{alg: adafgl-step1}
\begin{algorithmic}[1] 
    \FOR {each communication round $t = 1, ..., T$}
    \FOR {\textbf{parallel} each client $n = 1, ..., N$}
    \FOR {each local model training epoch $e = 1, ..., E$}
    \STATE Update local model weight according to Eq.~(\ref{eq:localtrain});\\
    \ENDFOR
    \ENDFOR 
    \STATE Execute server-side model aggregation based on Eq.~(\ref{eq:fedavg});\\
    \ENDFOR
    \STATE Broadcast the federated knowledge extractor to clients;\\
    \FOR {\textbf{parallel} each client $i = 1, ..., N$}
        \STATE Calculate the federated knowledge-guided probability propagation matrix $\widetilde{\mathbf{P}}_{i}$ according to Eq.(\ref{eq:correct_local_topology}) and Eq.~(\ref{eq:scaleA});\\
    \ENDFOR
\end{algorithmic}
\end{algorithm}

\begin{algorithm}[t]
\caption{AdaFGL-Adaptive Personalized Propagation} 
\label{alg: adafgl-step2}
\begin{algorithmic}[1] 
    \FOR {\textbf{parallel} each client $i = 1, ..., N$}
    \STATE Execute $K$-step Non-param LP process by Eq.(\ref{eq:labelpropagation}) to obtain the topology-aware label distribution $\hat{\mathbf{Y}}$;\\
    \STATE Calculate the homophily confidence score HCS according to the Eq.~(\ref{eq:hcs});\\
       \FOR {each local model training epoch $e = 1, ..., E$}
        \STATE Execute homophilous propagation function $f_{ho}$ to obtain $\hat{\mathbf{Y}}_{ho}$ by Eq.~(\ref{eq:homoensemble});\\
        \STATE Execute heterophilous propagation function $f_{he}$ to obtain $\hat{\mathbf{Y}}_{he}$ by Eq.~(\ref{eq:heteensmble});
        \STATE Calculate the local prediction $\hat{\mathbf{Y}}$ by Eq.~(\ref{eq:finalpredictions});\\
    \ENDFOR
    \STATE Update client-independent homophilous weights $\Theta_{Ho}$ and heterophilous weights $\Theta_{He}$ according to Eq.~(\ref{eq:loss});
    \ENDFOR 
\end{algorithmic}
\end{algorithm}

\subsection{Local Personalized Propagation}
\label{sec: local personalized propagation}
    Based on the federated knowledge-guided probability propagation matrix $\widetilde{\mathbf{P}}$, each client performs personalized training, which involves both homophilous and heterophilous propagation.
    To begin with, AdaFGL performs $k$-step federated knowledge-guided smoothing via $\widetilde{\mathbf{P}}$ to obtain knowledge embeddings $\widetilde{\mathbf{H}}$, which is formally expressed as
    \begin{equation}
    \label{eq:propadj_feature_propagtion}
      \begin{aligned}
        \widetilde{\mathbf{X}}^{(k)}=\operatorname{GraphOperator}\left(\widetilde{\mathbf{P}}\right)^k \mathbf{X}^{(0)},\;\;\;\;\;\;\;\;\;\;\;\;\;\;\\
        \widetilde{\mathbf{H}} = \operatorname{MessageUpdater}\left(\widetilde{\mathbf{X}}^{(1)},\dots,\widetilde{\mathbf{X}}^{(k)}, \Theta_{knowledge}\right),
      \end{aligned}
    \end{equation}
    where $\operatorname{GraphOperator}(\cdot)$ represents the topological smoothing operator in feature propagation, and we use symmetric normalized adjacency shown in Eq.~(\ref{eq:gcn}) as default.
    After $k$-step feature propagation, we correspondingly get a list of propagated features $[\widetilde{\mathbf{X}}^{(1)},\dots,\widetilde{\mathbf{X}}^{(k)}]$.
    Based on this, we utilize $\operatorname{MessageUpdater}(\cdot)$ with trainable parameters $\Theta_{knowledge}$ to achieve federated knowledge-guided learning.
    In our implementation, we use a multi-layer perceptron (MLP) to learn concatenated propagation features $[\widetilde{\mathbf{X}}^{(1)},||\dots||,\widetilde{\mathbf{X}}^{(k)}]$.
    
\subsubsection{Homophilous Propagation}   
\label{sec: homophilous propagation}

    This module abides by homophily assumption and leverages the reliable knowledge embeddings $\widetilde{\mathbf{H}}$ generated by the federated knowledge extractor. 
    {Such a design is based on the fact that feature propagation satisfying homophily imposes a positive impact on predictions~\cite{wang2020gcnlpa,huang2020cands,dai2022lwgcn}, which has been proven by graph denoise perspective.
    Meanwhile, $\widetilde{\mathbf{H}}$ performs well in homophily, as validated in Fig.~\ref{fig:motivation_exp} and Fig.~\ref{fig:motivation_optima}. 
    Therefore, AdaFGL achieves local smoothing under the supervision of knowledge embeddings. 
    }

    \underline{\emph{Feature Propagation}}.
    Due to the homophily topology in the local subgraph, the local predictions generated by the federated knowledge extractor have high confidence. 
    Therefore, the optimized initial probability propagation matrix $\widetilde{\mathbf{P}}$ can well generalize the local topology.
    Based on this, we directly obtain $\widetilde{\mathbf{H}}$ by Eq.~(\ref{eq:propadj_feature_propagtion}) without additional computation.

    \underline{\emph{Knowledge Preserving}}.
    As we have clarified above knowledge embeddings have a positive impact on local predictions.
    We introduce reliable supervised information based on the federated knowledge extractor.
    Specifically, we propose to define the following knowledge preserving loss function as
    \begin{equation}
    \label{eq:kploss}
      \begin{aligned}
        &\mathcal{L}_{knowledge}= \left|\left|\widetilde{\mathbf{H}}-\hat{\mathbf{X}}\right|\right|_F. 
      \end{aligned}
    \end{equation}
    
    \underline{\emph{Comprehensive Prediction}}.
    Considering the efficiency and performance issues, we generate the homophilous prediction via knowledge embeddings and local embeddings
    \begin{equation}
    \label{eq:homoensemble}
      \begin{aligned}
            \hat{\mathbf{Y}}_{ho} = \left(\mathrm{Softmax}\left(\widetilde{\mathbf{H}}\right)+\hat{\mathbf{P}}\right) / 2.
      \end{aligned}
    \end{equation}

\subsubsection{Heterophilous Propagation}
\label{sec: heterophilous propagation}
    To overcome the limitation of heterophilous topology on the local performance shown in Fig.~\ref{fig:motivation_exp}, we achieve personalized learning via the following intuitive perspectives: 
    (a) Topology-independent feature embedding; 
    (b) Global-dependent node embedding; 
    (c) Propagation-dependent message embedding. 
    In the following content, we will present the motivations and details of them.

    \underline{\emph{Topology-independent Feature Embedding}}.
    {Since the heterophilous topology brings uncontrollable results in the naive feature propagation as shown in Sec.~\ref{sec: prior gnns in central graph learning}, a natural idea is to implement topology-independent feature embedding, which focuses on heterophilous subgraph attribute mining. 
    Then, we can use this feature embedding to generate comprehensive predictions and to serve as a component of the framework for facilitating overall improvement. 
    Some recent research~\cite{zhu2021cpgnn,2021linkx,xu2022hpgmn} has demonstrated its effectiveness from the collaborative optimization perspective.
    The above process with MLP parameters $\Theta_{feature}$ is formally expressed as}
    \begin{equation}
    \label{eq:hete_ori_feature}
      \begin{aligned}
            &\mathbf{\mathbf{H}}_f = f(\mathbf{X},\Theta_{feature}).
      \end{aligned}
    \end{equation}
    
    \underline{\emph{Global-dependent Nodes Embedding}}.
    {From the attention representation perspective, recent heterophilous GNN~\cite{NEURIPS2021_5857d68c_ugcn,lei2022evennet,song2023ordergnn} theoretically prove that incorporating suitable high-order topology can improve the predictions on heterophily.} 
    In Step 1, the optimized probability propagation matrix $\widetilde{\mathbf{P}}$ to imply higher-order dependencies of each node. 
    Thus, we can directly obtain $\widetilde{\mathbf{H}}$ using Eq.~(\ref{eq:propadj_feature_propagtion}).
    However, we omit the ”Knowledge Preserving” step due to the unreliable knowledge embeddings caused by the local heterophilous topology.

    \underline{\emph{Learnable Message-passing Embedding}}.
    To mitigate the negative impact of heterophilous topology, we propose an end-to-end learnable mechanism for modeling the message passing, including positive and negative messages.
    {This idea is motivated by the success of message modeling mechanisms in heterophilous graphs~\cite{yan2021ggcn,2022glognn,liu2023simga}, which theoretically analyze the relations between message properties and topology.
    Notably, our personalized local message modeling mechanism offers improved practical performance.
    This is achieved by the optimized $\widetilde{\mathbf{P}}$ to capture the global dependency of the current node based on probability differences.
    Based on this, we then model the positive and negative impacts of the messages through local training, which can be formally represented as}
    \begin{equation}
    \label{eq:hetefeatureproprgation}
      \begin{aligned}
        \mathbf{H}_m^{(l)} = f\left(\mathbf{H}_m^{(l-1)},\Theta_{message}\right),\;\;\;\;\;\;\;\;\;\;\;\;\;\;\;\;\;\;\\
        \widetilde{\mathbf{P}}^{(l)} = \beta\widetilde{\mathbf{P}}^{(l-1)} +(1-\beta)\left( \mathbf{H}_m^{(l)}({\mathbf{H}_m^{(l)}})^\mathrm{T}\right),\;\;\;\;\;\;\;\;\;\;\;\;\\
        \mathbf{H}_{pos}^{(l)} = \mathrm{PoSign}\left(\widetilde{\mathbf{P}}^{(l)}\right)\mathbf{H}_m^{(l)}, \;\mathbf{H}_{neg}^{(l)} = \mathrm{NeSign}\left(\widetilde{\mathbf{P}}^{(l)}\right)\mathbf{H}^{(l)}_m,
      \end{aligned}
    \end{equation}
    where ${\mathbf{H}}_p^{(0)} = \widetilde{\mathbf{H}}, \widetilde{\mathbf{P}}^{(0)}=\widetilde{\mathbf{P}}$, $\mathrm{PoSign(\cdot)}$ and $\mathrm{NeSign(\cdot)}$ represent the message modeling functions, which can be replaced by any reasonable activation function.
    We use $f(\cdot)$ to represent $l$-layer MLP parametered by $\Theta_{message}$.
    After that, we combine the current embedding, positive embedding, and negative embedding to generate $\mathbf{H}^{(l+1)}_m$, which is defined as
    \begin{equation}
    \label{eq:hetefeaturecombination}
      \begin{aligned}
        \mathbf{H}^{(l+1)}_m = \mathbf{H}^{(l)}_m + \mathbf{H}_{pos}^{(l)} - \mathbf{H}_{neg}^{(l)}.
      \end{aligned}
    \end{equation}

    \underline{\emph{Comprehensive Prediction}}.
   Based on the above three perspectives for performing local personalized propagation, we generate the local heterophilous prediction as follows for considering practical performance and efficiency
    \begin{equation}
    \label{eq:heteensmble}
      \begin{aligned}
            \hat{\mathbf{Y}}_{he} = \left(\mathrm{Softmax}(\mathbf{H}_f)+\mathrm{Softmax}(\widetilde{\mathbf{H}}) + \mathrm{Softmax}(\mathbf{H}_m)\right) / 3.
      \end{aligned}
    \end{equation}
    
\subsubsection{Loss Function}
    In federated semi-supervised single-label node classification, we formulate the overall loss function in AdaFGL to achieve end-to-end training on each client
    \begin{equation}
    \label{eq:loss}
      \begin{aligned}
        \mathcal{L} = \mathcal{L}_{CE} + \mathcal{L}_{knowledge},
      \end{aligned}
    \end{equation}
    where $\mathcal{L}_{knowledge}$ represents the knowledge preserving loss in Eq.~(\ref{eq:kploss}), $\mathcal{L}_{CE}$ is the Cross-Entropy (CE) measurement between the predicted softmax outputs and the one-hot ground-truth label distributions shown in Eq.~(\ref{eq:localtrain}).
    Remarkably, in Step 1 of standard federated collaborative training, each client performs local training using only $\mathcal{L}_{CE}$. 
    In Step 2, the training process is further optimized by incorporating $\mathcal{L}_{knowledge}$ generated by the federated knowledge extractor.

\subsection{Homophily Confidence Score}
\label{sec: homophily confidence score}
    To enable adaptive mechanisms, AdaFGL utilizes a $K$-step Non-parametric Label Propagation (Non-param LP) approach to capture deep structural information.
    This procedure establishes connections between the current node and its $K$-hop neighbors based on the sparse matrix multiplication.
    Subsequently, AdaFGL calculates the homophily confidence score (HCS), which quantifies the degree of homophily in each client.
    Finally, the HCS is used to combine the homophilous and heterophilous propagation modules, generating final predictions.
    Remarkably, the above processes do not involve any learning over the local subgraph and maintain high computation efficiency.
    
    To implement the Non-param LP, the labeled nodes are initialized as $\mathbf{y}_i^0 = \mathbf{y}_i, \forall i \in V_L$, and the unlabeled nodes are denoted as $\mathbf{y}_j^0 = (\frac{1}{|Y|},\dots,\frac{1}{|Y|}), \forall j \in V_U$.
    Then, the $K$-step Non-param LP is expressed as
    \begin{equation}
    \label{eq:labelpropagation}
      \begin{aligned}
        \hat{\mathbf{Y}}_u^K = \kappa\hat{\mathbf{Y}}_u^0 + (1-\kappa)\sum_{v\in\mathcal{N}_u}\frac{1}{\sqrt{\tilde{d}_v\tilde{d}_u}}\hat{\mathbf{Y}}_v^{K-1}.
      \end{aligned}
    \end{equation}
    We adopt the approximate calculation method for the personalized PageRank~\cite{2019appnp}, where $\mathcal{N}_v$ denotes the one-hop neighbors of node $v$. 
    Meanwhile, we set $\kappa=0.5$ and $K=5$ by default to capture deep structural information.
    Due to the small-world phenomenon, we aim to traverse as many nodes as possible within the subgraph through such settings.
    Subsequently, we introduce the HCS to measure the homophily in the local subgraph, even in the absence of known node labels.
    \begin{definition}
    (Homophily Confidence Score, HCS)
    Given the training label matrix $\mathbf{Y}^{train}$, the set of nodes $V_L,V_U$, the mask function $\mathrm{mask}(\cdot)$, execute $K$-step Non-param LP with $\mathbf{y}_i^0 = \mathbf{y}_i, \forall i \in V_L / \mathrm{mask} (V_L)$ labeled initialization and $\mathbf{y}_j^0 = (\frac{1}{|Y|},\dots,\frac{1}{|Y|}), \forall j \in V_U \cup \mathrm{mask}(V_L)$ unlabeled initialization to obtain $[\hat{\mathbf{Y}}^1,\dots,\hat{\mathbf{Y}}^K]$. 
    We calculate the accuracy of the masked training nodes as the homophily confidence score
    \begin{equation}
    \label{eq:hcs}
      \begin{aligned}
        \operatorname{HCS}=\frac{\sum_{i\in\mathrm{mask}(V_L)}\mathrm{Count}(\hat{\mathbf{y}}_i^K==\mathbf{y}_i^{train})}{|\mathrm{mask}(V_L)|},
      \end{aligned}
    \end{equation}
    where $\mathrm{Count}(\cdot)$ is used to calculate the number of predicted correct nodes, and $|\mathrm{mask}(V_L)|$ represents the number of masked nodes.
    The masking probability is $0.5$ by default. 
    \end{definition}
    Finally, we leverage the topology-awareness HCS to adaptively combine the outputs of the homophilous propagation module and the heterophilous propagation module, generating the final predictions as follows
    \begin{equation}
    \label{eq:finalpredictions}
      \begin{aligned}
        \hat{\mathbf{Y}} = \operatorname{HCS}\hat{\mathbf{Y}}_{ho} + (1-\operatorname{HCS})\hat{\mathbf{Y}}_{he}.
      \end{aligned}
    \end{equation}

\section{Experiments}
    In this section, we present a comprehensive evaluation of our proposed AdaFGL. 
    We first introduce 12 benchmark datasets commonly used in central graph learning, including homophily and heterophily. 
    We then offer detailed descriptions of two distributed subgraph simulation methods: community split and structure Non-iid split. 
    We also present a comparison of baselines and detailed experiment settings, including state-of-the-art FGL approaches and federated implementations of representative GNNs designed for homophily and heterophily. 
    After that, we aim to address the following questions:
    \textbf{Q1}: Can AdaFGL achieve {better} predictive performance than state-of-the-art baselines under two federated settings?
    \textbf{Q2}: If AdaFGL is effective, what contributes to its performance gain?
    \textbf{Q3}: What are the advantages of AdaFGL as a new paradigm in the FGL?
    \textbf{Q4}: How does AdaFGL perform under sparse settings, such as low label/edge rate, missing features in multi-client subgraphs, and low client participation rate?

\begin{table*}[t]
\caption{The statistical information of the experimental datasets, "E.Homo" represents edge homophily mentioned in the Sec.~\ref{sec: Preliminaries}.
}
\vspace{-0.2cm}
\footnotesize 
\label{tab: datasets}
\resizebox{\linewidth}{26mm}{
\setlength{\tabcolsep}{3mm}{
\begin{tabular}{ccccccccc}
\midrule[0.3pt]
Datasets   & \#Nodes & \#Features & \#Edges    & \#Classes & \#Train/Val/Test & \#E.Homo & \#Task      & Description           \\ \midrule[0.3pt]
Cora       & 2,708   & 1,433      & 5,429      & 7         & 20\%/40\%/40\%   & 0.810     & Transductive & citation network      \\
CiteSeer   & 3,327   & 3,703      & 4,732      & 6         & 20\%/40\%/40\%   & 0.736     & Transductive & citation network      \\
PubMed     & 19,717  & 500        & 44,338     & 3         & 20\%/40\%/40\%   & 0.802     & Transductive & citation network      \\
Computer   & 13,381  & 767        & 245,778    & 10        & 20\%/40\%/40\%   & 0.777     & Transductive & co-purchase network   \\
Physics    & 34,493  & 8,415      & 247,962    & 5         & 20\%/40\%/40\%   & 0.931     & Transductive & co-authorship network \\ \midrule[0.3pt]
Chameleon  & 2,277   & 2,325      & 36,101     & 5         & 60\%/20\%/20\%   & 0.234     & Transductive & wiki pages network    \\
Squirrel   & 5,201   & 2,089      & 216,933    & 5         & 60\%/20\%/20\%   & 0.223     & Transductive & wiki pages network    \\
Actor      & 7,600   & 931        & 29,926     & 5         & 60\%/20\%/20\%   & 0.216     & Transductive & movie network         \\
Penn94     & 41,554  & 5          & 1,362,229  & 2         & 60\%/20\%/20\%   & 0.470     & Transductive & dating network        \\
arxiv-year & 169,343 & 128        & 1,166,243  & 5         & 60\%/20\%/20\%   & 0.222     & Transductive & publish network       \\ \midrule[0.3pt]
Reddit     & 89,250  & 500        & 899,756    & 7         & 44k/22k/22k      & 0.756     & Inductive    & image network         \\
Flickr     & 232,965 & 602        & 11,606,919 & 41        & 155k/23k/54k     & 0.319     & Inductive    & social network        \\ \midrule[0.3pt]
\end{tabular}
}}
\vspace{-0.2cm}
\end{table*}

\begin{table*}[t]
\setlength{\abovecaptionskip}{0.3cm}
\caption{Transductive performance under two simulation strategies.
The best result is \textbf{bold}.
The second result is \ul{underlined}.
}
\vspace{-0.25cm}
\footnotesize 
\label{tab: trans_cmp}
\resizebox{\linewidth}{43mm}{
\setlength{\tabcolsep}{1.5mm}{
\begin{tabular}{cc|cccccccccc}
\midrule[0.3pt]
Simulation                                                                    & Method     & Cora              & CiteSeer          & PubMed            & Computer          & Physics           & Chameleon         & Squirrel          & Actor             & Penn94            & arxiv-year        \\ \midrule[0.3pt]
\multirow{11}{*}{\begin{tabular}[c]{@{}c@{}}Community\\ Split\end{tabular}}   & GCN    & 79.4±0.4          & 68.5±0.3          & 85.9±0.1          & 80.5±0.4          & 88.7±0.5          & 60.5±2.1          & 46.9±1.4          & 28.8±1.1          & 79.6±0.7          & 46.7±0.9          \\
                                                                              & GCNII  & 80.7±0.6          & 69.8±0.4          & 86.1±0.2          & 81.0±0.7          & 89.3±0.7          & 59.8±2.6          & 40.7±2.0          & 30.4±1.6          & 77.4±0.9          & 46.5±1.1          \\
                                                                              & GAMLP  & 80.5±0.7          & 70.2±0.6          & 86.7±0.2          & 80.8±0.5          & 89.6±0.8          & 58.4±2.9          & 41.6±2.3          & 29.2±2.2          & 77.0±1.0          & 46.2±1.2          \\
                                                                              & GGCN   & 79.6±0.6          & 68.4±0.7          & 86.0±0.3          & 80.8±0.7          & 88.5±0.7          & 65.3±2.6          & 50.7±2.7          & {\ul 33.9±1.8}    & 78.8±1.3          & 48.9±1.2          \\               & GloGNN & 79.7±0.6          & 68.9±0.6          & 85.8±0.3          & 80.9±0.6          & 88.9±0.8          & {\ul 65.8±2.8}    & {\ul 51.8±2.1}    & 33.5±1.6          & 80.4±1.1          & {\ul 49.3±1.3}    \\               
                                                                              & GPRGNN & 79.2±0.5          & 68.7±0.5          & 85.9±0.2          & 80.7±0.5          & 88.5±0.8          & 62.6±2.2          & 46.8±2.5          & 32.4±1.4    & 78.3±1.1          & 45.8±1.0          \\
                                                                              & FedGL     & 80.2±0.8          & 69.7±0.8          & 86.5±0.3          & 81.3±0.9          & 89.4±1.1          & 62.1±3.4          & 48.9±2.4          & 32.0±1.9          & {\ul 81.3±1.6}    & 48.1±1.7          \\
                                                                              & GCFL+     & 79.8±0.2          & 69.4±0.3    & 85.7±0.1    & 79.8±0.2    & 88.6±0.2    & 58.4±1.9          & 42.8±0.9          & 29.6±0.7    & 79.2±0.5          & 46.3±0.4          \\
                                                                              & FedSage+  & 81.3±0.9          & {\ul 71.6±1.1}    & {\ul 86.8±0.5}    & {\ul 82.1±1.2}    & 90.0±1.5          & 59.4±3.9          & 42.7±3.0          & 29.8±2.6          & 80.0±1.8          & 44.9±2.1          \\
                                                                              & FED-PUB   & {\ul 81.5±0.3}    & 70.8±0.3          & 86.5±0.1          & 81.5±0.3          & {\ul 90.3±0.3}    & 59.3±1.8          & 43.3±1.2          & 29.2±1.2          & 79.8±0.6          & 46.0±0.6          \\
                                                                              & AdaFGL    & \textbf{82.9±0.5} & \textbf{72.5±0.6} & \textbf{88.4±0.2} & \textbf{83.6±0.4} & \textbf{90.8±0.6} & \textbf{68.9±2.1} & \textbf{56.1±1.5} & \textbf{35.9±1.5} & \textbf{83.5±0.7} & \textbf{51.2±0.8} \\ \midrule[0.3pt]
\multirow{11}{*}{\begin{tabular}[c]{@{}c@{}}Structure\\ Non-iid\end{tabular}} & GCN    & 70.6±0.6          & 63.4±0.5          & 82.3±0.2          & 72.3±0.5          & 83.2±0.4          & 63.8±2.0          & 51.2±1.8          & 35.4±1.2          & 82.3±0.6          & 49.6±0.8          \\
                                                                              & GCNII  & 70.8±0.7          & 63.9±0.7          & 82.6±0.3          & 72.1±0.6          & 84.1±0.8          & 64.7±2.8          & 48.5±2.2          & 37.2±1.5          & 78.5±0.7          & 47.8±1.0          \\
                                                                              & GAMLP  & 70.2±0.9          & 63.2±0.8          & 82.4±0.4          & 72.6±0.5          & 83.8±0.7          & 60.3±2.5          & 44.9±2.6          & 36.8±1.8          & 79.6±1.2          & 48.4±1.3          \\
                                                                              & GGCN   & 74.5±0.9          & {\ul 65.7±1.1}    & 84.8±0.4          & 75.3±0.6          & {\ul 86.3±0.6}    & 66.9±2.2          & 52.4±2.5          & 38.2±1.4    & 81.8±1.6          & 50.7±1.1          \\
                                                                              & GloGNN & {\ul 74.8±0.8}    & 65.4±1.2          & {\ul 85.1±0.3}    & {\ul 76.0±0.5}    & 85.8±0.8          & {\ul 67.4±2.5}    & {\ul 53.0±2.7}    & {\ul 39.4±2.0}    & 82.9±1.3          & {\ul 51.2±1.5}    \\
                                                                              & GPRGNN & 71.8±0.8          & 63.9±0.8    & 83.7±0.3          & 74.8±0.6          & 84.6±0.9    & 64.2±2.7          & 48.6±2.1          & 37.9±1.8          & 80.7±1.4          & 48.9±1.4          \\
                                                                              & FedGL     & 72.1±1.0          & 64.0±0.9          & 83.9±0.4          & 75.3±1.1          & 84.0±1.2          & 63.5±3.0          & 51.3±2.0          & 35.8±1.6          & {\ul 83.3±1.5}    & 49.5±1.9          \\
                                                                              & GCFL+     & 70.8±0.3          & 63.5±0.2          & 82.5±0.1          & 72.1±0.1          & 83.9±0.3          & 61.1±1.5          & 46.7±0.8          & 34.2±0.6          & 80.7±0.4          & 48.1±0.5          \\
                                                                              & FedSage+  & 72.6±1.1          & 64.5±1.2    & 84.2±0.6          & 74.6±1.4          & 85.7±1.3    & 62.4±3.5          & 45.1±3.2          & 33.7±2.8    & 81.0±2.2          & 47.3±1.8          \\
                                                                              & FED-PUB   & 71.3±0.5          & 63.2±0.5          & 82.4±0.1          & 72.9±0.2          & 83.4±0.2          & 61.6±1.6          & 48.2±1.4          & 33.5±1.1          & 80.0±0.8          & 47.4±0.7          \\
                                                                              & AdaFGL    & \textbf{80.4±0.8} & \textbf{69.3±0.8} & \textbf{87.7±0.3} & \textbf{80.8±0.3} & \textbf{87.5±0.4} & \textbf{70.8±2.2} & \textbf{60.4±1.7} & \textbf{42.7±1.3} & \textbf{85.2±1.0} & \textbf{56.8±1.0} \\ \midrule[0.3pt]
\end{tabular}
}}
\vspace{-0.4cm}
\end{table*}

\subsection{Experimental Setup}
\label{sec: experimental setup}
\noindent \textbf{Datasets.}
    We evaluate AdaFGL on 12 datasets, considering both transductive and inductive settings, as well as homophily and heterophily.
    For homophily, we perform experiments on three citation networks (Cora, Citeseer, PubMed)~\cite{Yang16cora}, user-item dataset (Computer), and coauthor dataset (Physics)~\cite{shchur2018amazon_datasets} under transductive settings. 
    Additionally, we conduct experiments on image dataset~(Flickr)~\cite{zeng2019graphsaint} under inductive settings.
    Regarding heterophily, we conduct experiments on two wiki networks (Chameleon, Squirrel)~\cite{pei2020geomgcn}, movie network (Actor)~\cite{pei2020geomgcn}, dating network (Penn94)~\cite{2021linkx}, and publish network (arxiv-year)~\cite{2021linkx} under transductive settings. 
    Moreover, we perform experiments on social network (Reddit)~\cite{zeng2019graphsaint} under inductive settings.
    For more statistical information, please refer to Table~\ref{tab: datasets}.

    Based on the above global graph, we apply two data simulation strategies in our experiments. 
    The community split approach is commonly employed in recent FGL studies~\cite{zhang2021fedsage,WangFedScope_22_fsg,baek2022fedpub}. 
    The structure Non-iid split we proposed serves as a new benchmark to bridge the research and application phases as we introduce in Sec.~\ref{sec: introduction}.
    To implement community split, we initially apply the Louvain algorithm~\cite{blondel2008louvain} on the global graph and assign communities to different clients following the nodes average principle. 
    This ensures a relatively uniform distribution of the number of nodes among the clients.
    Regarding the structure Non-iid split, we employ the Metis algorithm~\cite{karypis1998metis} to assign subgraphs based on the number of clients. 
    In detail, we propose random/meta-injection to introduce additional edges and achieve multi-client topology variances.
    {Specifically, we first perform a binary selection with $p_s=0.5$ for each subgraph to perform homogeneous (random-injection) or heterogeneous (random/meta-injection) structural injection.
    For the random-injection, we first generate increasing edges based on half (sampling ratio is 50\%) of the original edges and then randomly select non-connected node pairs for heterophilous perturbations or homophilous augmentation.
    For meta-injection generated by Metattack~\cite{zugner_adversarial_2019_metaattck}, we budget the attack as 0.2 of total edges in each dataset and only enhance heterophily in homophilous subgraphs since its unscrutinized injection in heterophilous subgraphs leads to performance degradation, which conflicts with intentions to enhance homophily to stir topology variance. 
    Consequently, we provide diverse evaluations of AdaFGL in different scenarios with a comprehensive analysis.}

\noindent \textbf{Baselines.}
    For federated implementation of GNNs, we compare AdaFGL with homophilous GCN~\cite{kipf2016gcn}, GCNII~\cite{chen2020gcnii}, GAMLP~\cite{gamlp} and heterophilous GPRGNN~\cite{chien2021gprgnn}, GGCN~\cite{yan2021ggcn}, GloGNN~\cite{2022glognn}.
    For FGL studies, we conduct comparisons on recently proposed FedGL~\cite{chen2021fedgl}, GCFL+~\cite{xie2021gcfl}, FedSage+~\cite{zhang2021fedsage}, and FED-PUB~\cite{baek2022fedpub}.
    The results we present are calculated by 10 runs.
    In default, we adopt the random-injection technique in the structure Non-iid split and apply a 10-client split since random-injection technique is user-friendly and serves purposes of homophilous and heterophilous injections.

\noindent \textbf{Hyperparameters.}
    The hyperparameters are set based on the original paper if available.
    Otherwise, we perform an automatic hyperparameter search via the Optuna~\cite{akiba2019optuna}.
    For the percentage of selected augmented nodes and the number of generated neighbors, we conduct a grid search from $\{$0.01, 0.05, 0.1, 0.5$\}$ and $\{$2, 5, 10$\}$.
    We set the hidden dimension for all datasets to 64 and the number of local epochs to 5. 
    In addition, we perform 100 rounds for federated training.
    Notably, to ensure a fair comparison, we perform local corrections for all federated implementations of GNNs {to} achieve maximum convergence for each client.
    For our proposed AdaFGL, the coefficient of topology optimization ($\alpha$) and learnable propagation rules ($\beta$) are explored within the ranges of 0 to 1.


\noindent \textbf{Experiment Environment.}
    The experimental machine with Intel(R) Xeon(R) Gold 6230R CPU @ 2.10GHz, and NVIDIA GeForce RTX 3090 with 24GB memory and CUDA 11.8.
    The operating system is Ubuntu 18.04.6 with 216GB memory.

\begin{table}[t]
\setlength{\abovecaptionskip}{0.3cm}
\setlength{\belowcaptionskip}{0cm}
\caption{Inductive performance under two simulation strategies.
}
\vspace{-0.2cm}
\footnotesize
\label{tab: ins_cmp}
\resizebox{\linewidth}{30mm}{
\setlength{\tabcolsep}{3.2mm}{
\begin{tabular}{cc|cc}
\hline
Simulation                                                                   & Method   & Flickr              & Reddit              \\ \midrule[0.3pt]
\multirow{7}{*}{\begin{tabular}[c]{@{}c@{}}Community\\ Split\end{tabular}}   & GCNII    & 49.54±0.37          & 92.23±0.16          \\
                                                                             & GloGNN   & {\ul 50.33±0.25}    & 91.87±0.12          \\
                                                                             & FedGL    & 50.14±0.34          & 92.36±0.25          \\
                                                                             & GCFL+    & 48.47±0.17          & 91.83±0.09          \\
                                                                             & FedSage+ & 48.82±0.47          & 92.88±0.28          \\
                                                                             & FED-PUB  & 49.20±0.21          & {\ul 93.08±0.14}    \\
                                                                             & AdaFGL   & \textbf{52.48±0.35} & \textbf{94.75±0.23} \\ \midrule[0.3pt]
\multirow{7}{*}{\begin{tabular}[c]{@{}c@{}}Structure\\ Non-iid\end{tabular}} & GCNII    & 54.17±0.34          & 90.85±0.19          \\
                                                                             & GloGNN   & {\ul 54.89±0.16}    & 91.32±0.16          \\
                                                                             & FedGL    & 52.96±0.40          & 90.93±0.30          \\
                                                                             & GCFL+    & 49.55±0.13          & 90.26±0.11          \\
                                                                             & FedSage+ & 53.46±0.47          & {\ul 91.40±0.34}    \\
                                                                             & FED-PUB  & 51.74±0.18          & 91.32±0.19          \\
                                                                             & AdaFGL   & \textbf{57.65±0.28} & \textbf{93.55±0.20} \\ \midrule[0.3pt]
\end{tabular}
}}
\vspace{-0.4cm}
\end{table}

\begin{table}[t]
\setlength{\abovecaptionskip}{0.3cm}
\setlength{\belowcaptionskip}{-0.1cm}
\caption{Transductive performance under 2 injection strategies.
}
\vspace{-0.1cm}
\scriptsize
\label{tab: trans_2noniid_exp}
\resizebox{\linewidth}{13mm}{
\setlength{\tabcolsep}{2.4mm}{
\begin{tabular}{c|cccc}
\hline
\multirow{2}{*}{Method} & \multicolumn{2}{c}{Physics}           & \multicolumn{2}{c}{Penn94}            \\
                        & Random            & Meta              & Random            & Meta              \\ \hline
FedGL                   & 84.0±1.2          & 81.9±1.6          & {\ul 83.3±1.5}    & {\ul 80.6±1.4}    \\
GCFL+                   & 83.9±0.3          & 82.3±0.5          & 80.7±0.4          & 78.9±0.3          \\
FedSage+                & {\ul 85.8±1.3}    & {\ul 82.6±1.7}    & 81.0±2.2          & 78.2±1.9          \\
FED-PUB                 & 83.4±0.2          & 81.8±0.4          & 80.0±0.8          & 78.4±0.6          \\
AdaFGL                  & \textbf{87.5±0.4} & \textbf{86.4±0.6} & \textbf{85.2±1.0} & \textbf{82.4±0.8} \\ \hline
\end{tabular}
}}
\end{table}

\begin{table}[!h]
\setlength{\abovecaptionskip}{0.3cm}
\setlength{\belowcaptionskip}{-0cm}
\caption{Inductive performance under 2 injection strategies.
}
\vspace{-0.2cm}
\scriptsize
\label{tab: ins_2noniid_exp}
\resizebox{\linewidth}{13mm}{
\setlength{\tabcolsep}{2.4mm}{
\begin{tabular}{c|cccc}
\hline
\multirow{2}{*}{Method} & \multicolumn{2}{c}{Flickr}            & \multicolumn{2}{c}{Reddit}            \\
                        & Random            & Meta              & Random            & Meta              \\ \hline
FedGL                   & 53.0±0.4          & {\ul 49.1±0.3}    & 90.9±0.3          & 88.7±0.4          \\
GCFL+                   & 49.6±0.1          & 48.0±0.1          & 90.3±0.1          & 88.2±0.2          \\
FedSage+                & {\ul 53.5±0.5}    & 46.8±0.6          & {\ul 91.4±0.3}    & 89.8±0.5          \\
FED-PUB                 & 51.7±0.2          & 47.8±0.2          & 91.3±0.2          & {\ul 90.2±0.2}    \\
AdaFGL                  & \textbf{57.7±0.3} & \textbf{51.0±0.2} & \textbf{93.6±0.2} & \textbf{92.5±0.3} \\ \hline
\end{tabular}
}}
\vspace{-0.4cm}
\end{table}

\subsection{Performance Comparison}

\noindent \textbf{Transductive Performance.} 
    To answer \textbf{Q1}, we report the results in Table~\ref{tab: trans_cmp} and Table~\ref{tab: trans_2noniid_exp} to predict test nodes based on existing subgraphs. 
    We observe that AdaFGL outperforms all the baselines. 
    For instance, in the community split scenario, FedSage+ and FED-PUB fail to achieve consistent and competitive performance on homophilous datasets. 
    Moreover, existing FGL methods struggle to deliver satisfactory performance on heterophilous datasets like Chameleon and Squirrel due to their limited considerations of heterophily. 
    Such limitations are amplified by the enhanced topology heterogeneity from the Structure Non-iid split. 
    However, heterophilous GNNs like FedGGCN and FedGloGNN under federated settings with enhanced heterophily have unique advantages.
    In contrast, AdaFGL demonstrates a significant performance improvement compared to the most competitive baselines in both the community split and structure Non-iid split, achieving an average improvement of 3.27\% and 6.43\%, respectively.
    {
    Furthermore, we observe that AdaFGL shows limited improvement in weak global homophily CiteSeer and feature-rich Physics. 
    We attribute this to the challenging distributed topologies, which constrained the expressive power of the federated knowledge extractor and over-fitting problems.
    }

\begin{figure}[t]
	\centering
    \setlength{\abovecaptionskip}{0.2cm}
    \setlength{\belowcaptionskip}{-0.2cm}
  \includegraphics[width=\linewidth]{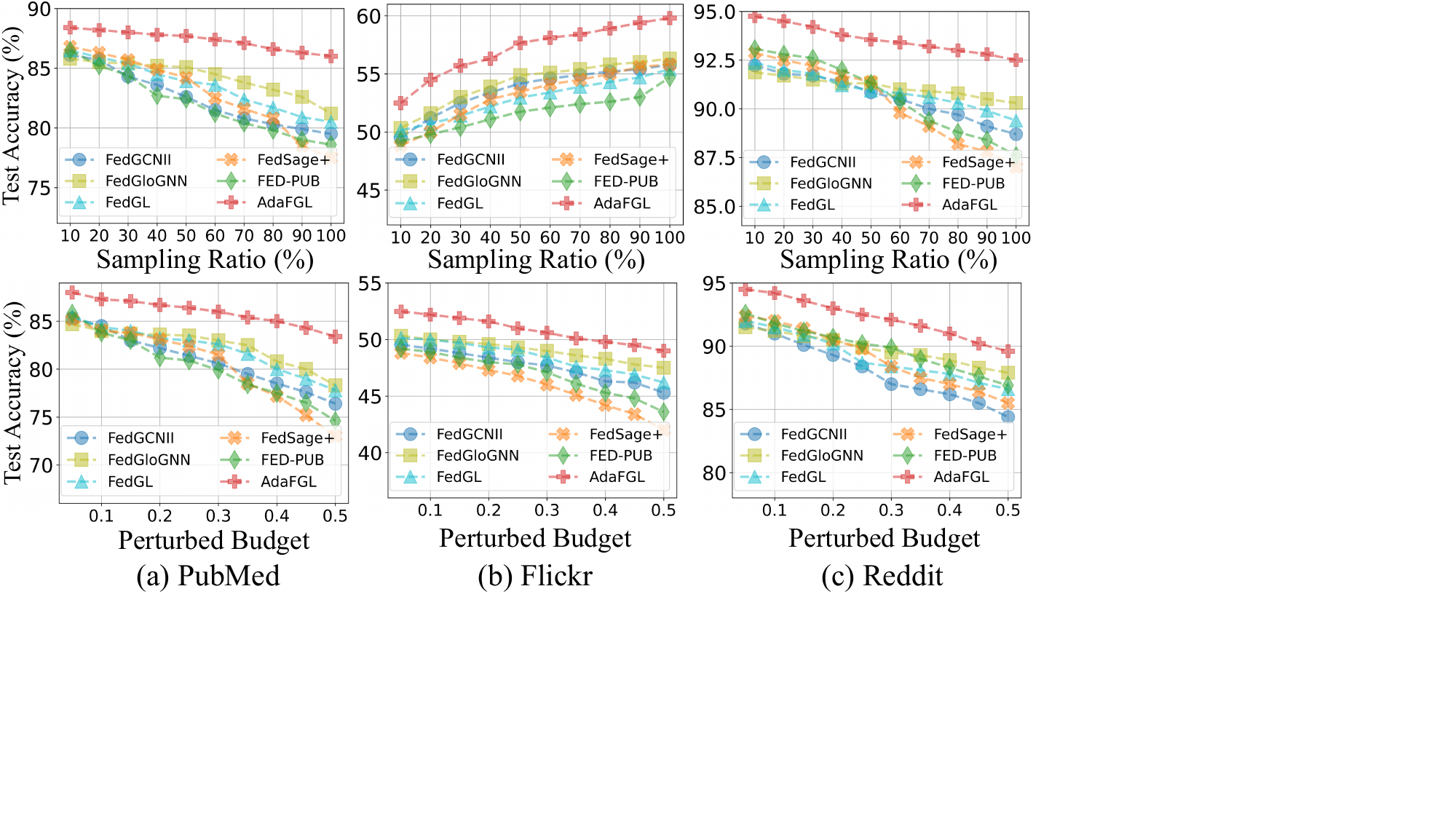}
  \caption{
   {Predictive performance under different topology heterogeneity.}}

  \label{fig:exp_hete}
\end{figure}

\noindent \textbf{Inductive Performance.} 
    Inductive learning aims to perform the node classification based on the partial training dataset within the same subgraph. 
    The experimental results in Table~\ref{tab: ins_cmp} and Table~\ref{tab: ins_2noniid_exp} consistently demonstrate the superior performance of AdaFGL compared to all baselines, under both community split and structure Non-iid split. 
    Notably, AdaFGL exhibits a significant lead over the state-of-the-art FedGloGNN on Flickr with a performance gap of more than 4.3\%. 
    Furthermore, AdaFGL achieves a remarkable improvement of over 2\% compared to the most competitive FedSage+ and FED-PUB on Reddit.
    The impressive performance of AdaFGL under the inductive setting highlights its ability to predict unseen nodes. 
    {
    Furthermore, AdaFGL's enhancements are constrained due to sparse feature dimensions and the limited number of labels in the Penn94. 
    Therefore, considering its transductive performance, we conclude that richer feature information supports AdaFGL's multi-module collaborative optimization, resulting in satisfactory predictive performance.}

\noindent{
\textbf{Different Topology Heterogeneity.}
    Inspired by deployment challenges from the topology heterogeneity in FGL mentioned in Sec.~\ref{sec: introduction}, we propose the structure Non-iid split to simulate a more generalized FGL setting. 
    In Fig.~\ref{fig:exp_hete}, we validate the model performance under various topology heterogeneity. 
    Notably, under the heterophilous Flickr, random-injection increases homophilous edges, consequently enhancing performance. 
    In contrast, meta-injection inspired by Metattack~\cite{zugner_adversarial_2019_metaattck} results in performance declines, making it only suitable to simulate structrue Non-iid in homophilous subgraphs to stir topology variance as described in Sec.~\ref{sec: experimental setup}.  
    Based on experimental results, we find that AdaFGL consistently maintains the best performance facing varying topology heterogeneity. 
    Moreover, compared to other baselines, AdaFGL exhibits a relatively gradual performance decrease.
}

\subsection{Ablation Study and Sensitivity Analysis}
    To answer \textbf{Q2}, we focus on three critical modules introduced in the local personalized training of AdaFGL: 
    (1) homophilous propagation (Homo.); 
    (2) heterophilous propagation (Hete.); 
    (3) adaptive mechanisms (Ada.). 
    For (1) and (2), the technical details about knowledge preserving (K.P.) in Homo., topology-independent feature embedding (T.F.), and learnable message-passing embedding (L.M.) in Hete. can be found in Sec.~\ref{sec: homophilous propagation} and Sec.~\ref{sec: heterophilous propagation}.
    Regarding (3), it involves local topology optimization (L.T.) and HCS, where L.T. is based on the federated knowledge extractor and HCS achieves the adaptive combination of homophilous and heterophilous outputs.

\begin{table}[t]
\setlength{\abovecaptionskip}{0.25cm}
\setlength{\belowcaptionskip}{-0.15cm}
\caption{Ablation study on homophilous datasets.
}
\vspace{-0.1cm}
\scriptsize
\label{tab: ab_exp_homo}
\resizebox{\linewidth}{20mm}{
\setlength{\tabcolsep}{1.5mm}{
\begin{tabular}{cc|cccc}
\midrule[0.3pt]
\multirow{2}{*}{Module}      & \multirow{2}{*}{Component} & \multicolumn{2}{c}{Computer}          & \multicolumn{2}{c}{Reddit}            \\
                             &                            & Com.         & Non-iid           & Com.         & Non-iid           \\ \midrule[0.3pt]
Homo.                    & w/o K.P.                   & 82.3±0.6          & 79.3±0.4          & 93.2±0.4          & 92.6±0.3          \\ \midrule[0.3pt]
\multirow{2}{*}{Hete.} & w/o T.F.                 & 83.2±0.4          & 78.5±0.5          & 94.3±0.2          & 93.1±0.3          \\
                             & w/o L.M.                   & 82.5±0.7          & 77.6±0.3          & 94.0±0.4          & 92.0±0.3          \\ \midrule[0.3pt]
\multirow{2}{*}{Ada.}    & w/o L.T.                   & 81.5±0.3          & 76.4±0.6          & 92.5±0.4          & 91.5±0.3          \\
                             & w/o HCS                    & 81.8±0.4          & 78.8±0.5          & 93.6±0.3          & 92.4±0.2          \\ \midrule[0.3pt]
AdaFGL                       & -                          & \textbf{83.6±0.4} & \textbf{80.8±0.3} & \textbf{94.8±0.2} & \textbf{93.6±0.2} \\ \midrule[0.3pt]
\end{tabular}
}}
\end{table}

\begin{table}[t]
\setlength{\abovecaptionskip}{0.25cm}
\setlength{\belowcaptionskip}{-0.15cm}
\caption{Ablation study on heterophilous datasets.
}
\vspace{-0.1cm}
\scriptsize
\label{tab: ab_exp_hete}
\resizebox{\linewidth}{20mm}{
\setlength{\tabcolsep}{1.5mm}{
\begin{tabular}{cc|cccc}
\midrule[0.3pt]
\multirow{2}{*}{Module}      & \multirow{2}{*}{Component} & \multicolumn{2}{c}{arxiv-year}        & \multicolumn{2}{c}{Flickr}            \\
                             &                            & Com.         & Non-iid           & Com.         & Non-iid           \\ \midrule[0.3pt]
Homo.                    & w/o K.P.                   & 51.0±1.1          & 56.1±1.2          & 52.2±0.5          & 57.2±0.3          \\ \midrule[0.3pt]
\multirow{2}{*}{Hete.} & w/o T.F.                   & 50.6±1.0          & 56.3±1.4          & 52.0±0.6          & 57.2±0.4          \\
                             & w/o L.M.                   & 48.7±0.8          & 54.3±1.1          & 49.8±0.4          & 55.4±0.3          \\ \midrule[0.3pt]
\multirow{2}{*}{Ada.}    & w/o L.T.                   & 49.4±0.8          & 55.0±0.9          & 51.2±0.6          & 56.5±0.4          \\
                             & w/o HCS                    & 50.5±0.9          & 55.9±1.1          & 51.7±0.5          & 57.0±0.4          \\ \midrule[0.3pt]
AdaFGL                       & -                          & \textbf{51.2±0.8} & \textbf{56.8±1.0} & \textbf{52.5±0.4} & \textbf{57.7±0.3} \\ \midrule[0.3pt]
\end{tabular}
}}
\vspace{-0.3cm}
\end{table}

\begin{figure*}[t]
  \includegraphics[width=\textwidth]{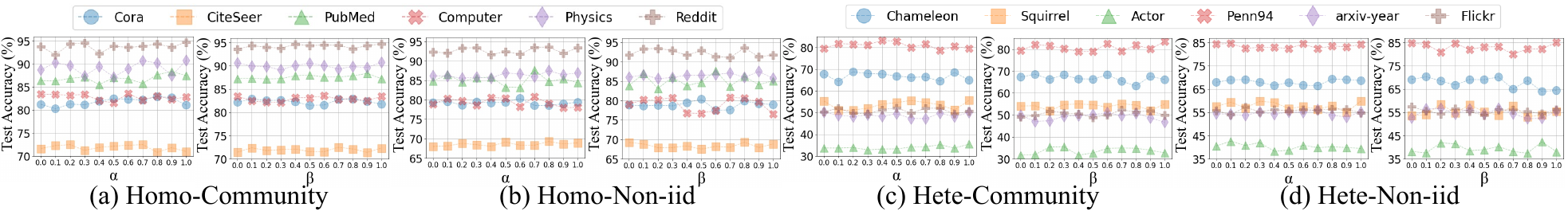}
  \vspace{-0.4cm}
  \caption{
    Hyperparameter sensitivity analysis on homophilous and heterophilous datasets with two data simulation strategies.}
  \label{fig:exp_hyperparameter}
\end{figure*}

\begin{figure*}[t]
  \includegraphics[width=\textwidth]{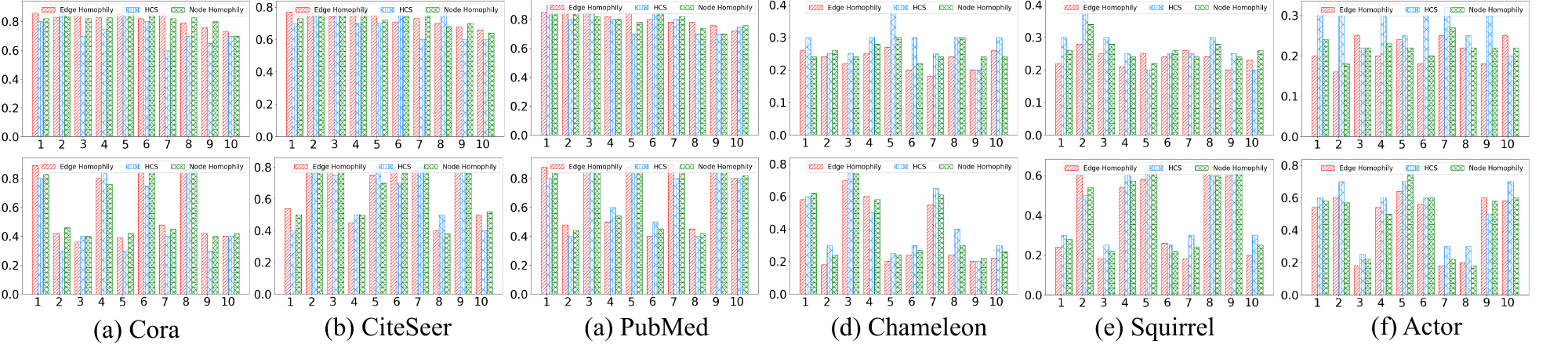}
  \vspace{-0.4cm}
  \caption{
    {Client-dependent HCS.
    The upper and lower represent the community and structure Non-iid split.
    The x-axis represents client id.}}
  \label{fig:exp_hcs}
\end{figure*}

\noindent \textbf{Personalized Propagation.}
    We evaluate the impact of incorporating decoupled components in the two above propagation modules.
    The experimental results in Table~\ref{tab: ab_exp_homo} demonstrate that the inclusion of K.P. in the Homo. significantly enhances the performance and stability of AdaFGL. 
    For example, in Computer, the accuracy improves from 82.3±0.6 to 83.6±0.4. 
    Conversely, removing the K.P. leads to a notable drop in performance due to over-fitting issues.
    Furthermore, the results in Table~\ref{tab: ab_exp_hete} indicate that both T.F. and L.M. play significant roles in the Hete.. 
    For instance, in arxiv-year, T.F. reduces the predictive error from 1.4 to 1.0, resulting in stable predictions by generating topology-independent embeddings. 
    Similarly, in Flickr, L.M. improves the performance from 55.4 to 57.7 by learning personalized propagation rules through optimized topology.
    Additionally, we conduct a sensitivity analysis of the hyperparameters associated with Homo. and Hete. as shown in Fig.~\ref{fig:exp_hyperparameter} with technical details provided in Eq.~(\ref{eq:correct_local_topology}) and Eq.~(\ref{eq:hetefeatureproprgation}).
    {Our experimental results demonstrate that larger $\alpha$ and $\beta$ are crucial for preserving the original topology in homophilous settings, while smaller $\alpha$ and $\beta$ are for optimizing propagation rules in heterophilous settings.}
    In Fig.~\ref{fig:exp_hyperparameter}(a), we observe that most datasets exhibit stable and improved performance when $\alpha$ is set to a larger value, indicating the presence of homophilous topologies. 
    Notably, the larger HCS value obtained by Eq.~(\ref{eq:hcs}) leads to a more dominant impact from Homo. in prediction, making the variation of $\beta$ less impactful on model performance.

\noindent \textbf{Adaptive Mechanisms.}
     In this part, we investigate the L.T. (Sec.~\ref{sec: federated knowledge extractor}) and HCS (Sec.~\ref{sec: homophily confidence score}), which influence the optimized topology and the final output. 
     The experimental results in Table~\ref{tab: ab_exp_homo} and Table~\ref{tab: ab_exp_hete} reveal the {superiority of combining L.T. and HCS.}
     This impressive performance gain lies in its ability to enable the personalized model to capture the local subgraph and generate multi-level predictions.
     Furthermore, the impacts of K.P., T.F., and L.M. are significant since their contributions are amplified adaptively through the L.T. and HCS, which further demonstrates the effectiveness of our adaptive mechanism.
     {
     To further support our claims, we visualize the HCS and the local subgraph homophily in Fig.~\ref{fig:exp_hcs}.
     In most cases, HCS is approximately equal to subgraph homophily, indicating the successful capture of local topology, which is used to guide personalized propagation and generate predictions.}

\begin{table}[t]
\setlength{\abovecaptionskip}{0.25cm}
\setlength{\belowcaptionskip}{-0.15cm}
\caption{A summary of recent FGL studies. 
}
\vspace{-0.1cm}
\scriptsize
\label{tab: fed_methods}
\resizebox{\linewidth}{23mm}{
\setlength{\tabcolsep}{1.2mm}{
\begin{tabular}{cc|c|c|c}
\hline
Methods  & Type & Communication                                                                                        & Server-side                                                                                                    & Client-side                                                                                                         \\ \hline
FedGL    & FedC & \begin{tabular}[c]{@{}c@{}}Model Param.\\ Node Pred.\\ Node Emb.\\ Graph Adj.\end{tabular}           & \begin{tabular}[c]{@{}c@{}}Model Agg.\\ Label Dis.\\ Graph Cons.\\ Broadcast Info.\end{tabular}         & \begin{tabular}[c]{@{}c@{}}Local Training\\ Pseudo Graph\\ Pseudo Prediction\\ Global Self-supervision\end{tabular} \\ \hline
GCFL+    & FedS & \begin{tabular}[c]{@{}c@{}}Model Param.\\ Model Grad.\end{tabular}                                   & \begin{tabular}[c]{@{}c@{}}Model Agg.\\ Grad. Clustering\end{tabular}                                       & Local Training                                                                                                      \\ \hline
FedSage+ & FedC & \begin{tabular}[c]{@{}c@{}}Model Param.\\ Node Emb.\\ NeighGen Param. \\ NeighGen Grad.\end{tabular} & \begin{tabular}[c]{@{}c@{}}Model Agg.\\ NeighGen Agg.\\ Broadcast Emb. \\ Broadcast Grad.\end{tabular} & \begin{tabular}[c]{@{}c@{}}Local Training\\ Data Augmentation\\ Send Embedding\\ Cross-client Gradient\end{tabular} \\ \hline
FED-PUB  & FedC & \begin{tabular}[c]{@{}c@{}}Model Param.\\ Model Mask\end{tabular}                                    & \begin{tabular}[c]{@{}c@{}} Broadcast Mask\\ Emb. Clustering \end{tabular}          & \begin{tabular}[c]{@{}c@{}}Local Training\\ Personalized Mask \end{tabular}                                          \\ \hline
\textbf{AdaFGL}   & FedC & Model Param.                                                                                         & Model Agg.                                                                                                     & \begin{tabular}[c]{@{}c@{}}Local Training\\ Personalized Prop. \end{tabular}                      \\ \hline
\end{tabular}
}}
\vspace{-0.15cm}
\end{table}

\begin{figure}[t]
	\centering
    \setlength{\abovecaptionskip}{0.2cm}
    \setlength{\belowcaptionskip}{-0.35cm}
  \includegraphics[width=\linewidth]{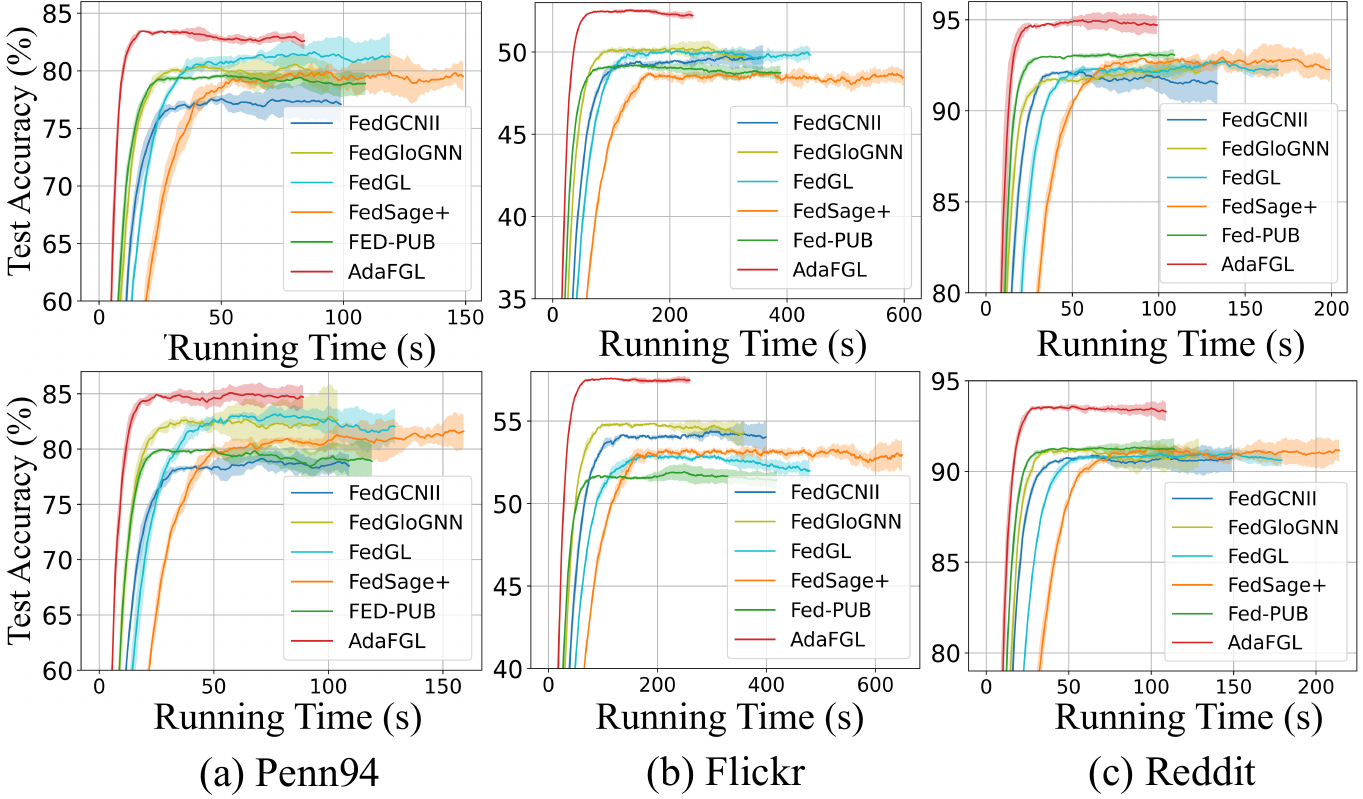}
  \caption{
    {Convergence curves of community split (upper) and structure Non-iid split (lower)
    The shadows are the result of 10 runs.}}

  \label{fig:exp_time_analysis}
\end{figure}

\begin{figure*}[t]
	\centering
    \setlength{\abovecaptionskip}{0.2cm}
    \setlength{\belowcaptionskip}{-0.2cm}
  \includegraphics[width=\textwidth]{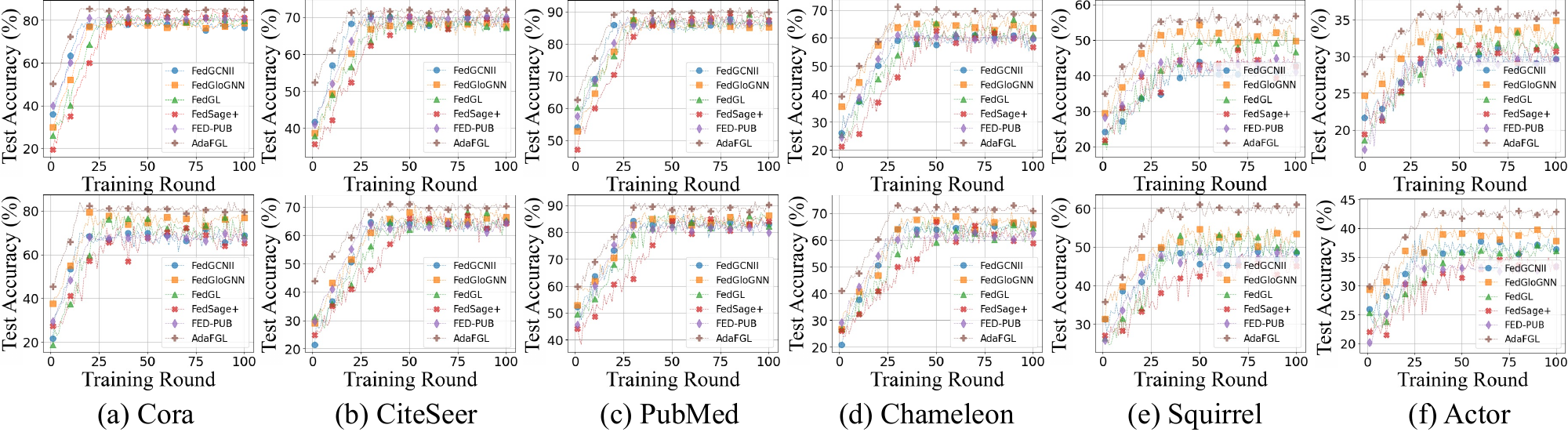}
  \caption{
    Convergence curves on the community split (upper) and structure Non-iid split (lower).
}
    \vspace{-0.1cm}
  \label{fig:exp_converge}
\end{figure*}

\subsection{Advantages of AdaFGL}
\noindent \textbf{New FGL Paradigm.}
    To address \textbf{Q3}, we first analyze the paradigm designs of existing FGL studies from a unified perspective and then highlight motivations behind AdaFGL.
    Recent FGL studies can be classified into two categories:
    (i)FedC: focusing on improved local model architectures and multi-client interactions on the client side;
    (ii)FedS: focusing on optimized model aggregation rules for federated training on the server side.
    We then review FGL methods in Table~\ref{tab: fed_methods}, revealing that existing methods often {present} significant communication overhead and privacy concerns. 
    For example, FedSage+ and FedGL require the exchange of node embeddings and other privacy-sensitive information.
    In contrast, AdaFGL addresses these challenges by maximizing the computational capacity of the local system while minimizing communication costs and the risk of privacy leakage through our decoupled two-step personalized design. 
    This aligns with the design of FED-PUB, which uses a model masking mechanism based on local training.
    However, AdaFGL transcends limitations of the original FGL approaches and introduces a new design paradigm, providing novel insights and advancements in the field.

\noindent \textbf{Training Efficiency.}
    Using federated knowledge combined with local subgraphs for personalized training, AdaFGL oversteps limits of communication overhead caused by device heterogeneity. 
    It enables efficient paralleled local propagation after the training of federated knowledge extractor. 
    The visualized converging process in Fig.~\ref{fig:exp_time_analysis} demonstrates that AdaFGL maintains high training efficiency based on the curve and shaded trends.
    Furthermore, in Fig.~\ref{fig:exp_converge}, we observe that AdaFGL presents higher initial performance compared to other methods and enables a faster and more stable convergence.
    For instance, in the structure Non-iid split for the Squirrel dataset, AdaFGL achieves nearly converged performance by the 25th epoch, and it remains stable throughout the subsequent training process.

\begin{figure}[t]
	\centering
    \setlength{\abovecaptionskip}{0.2cm}
    \setlength{\belowcaptionskip}{-0.15cm}
  \includegraphics[width=\linewidth]{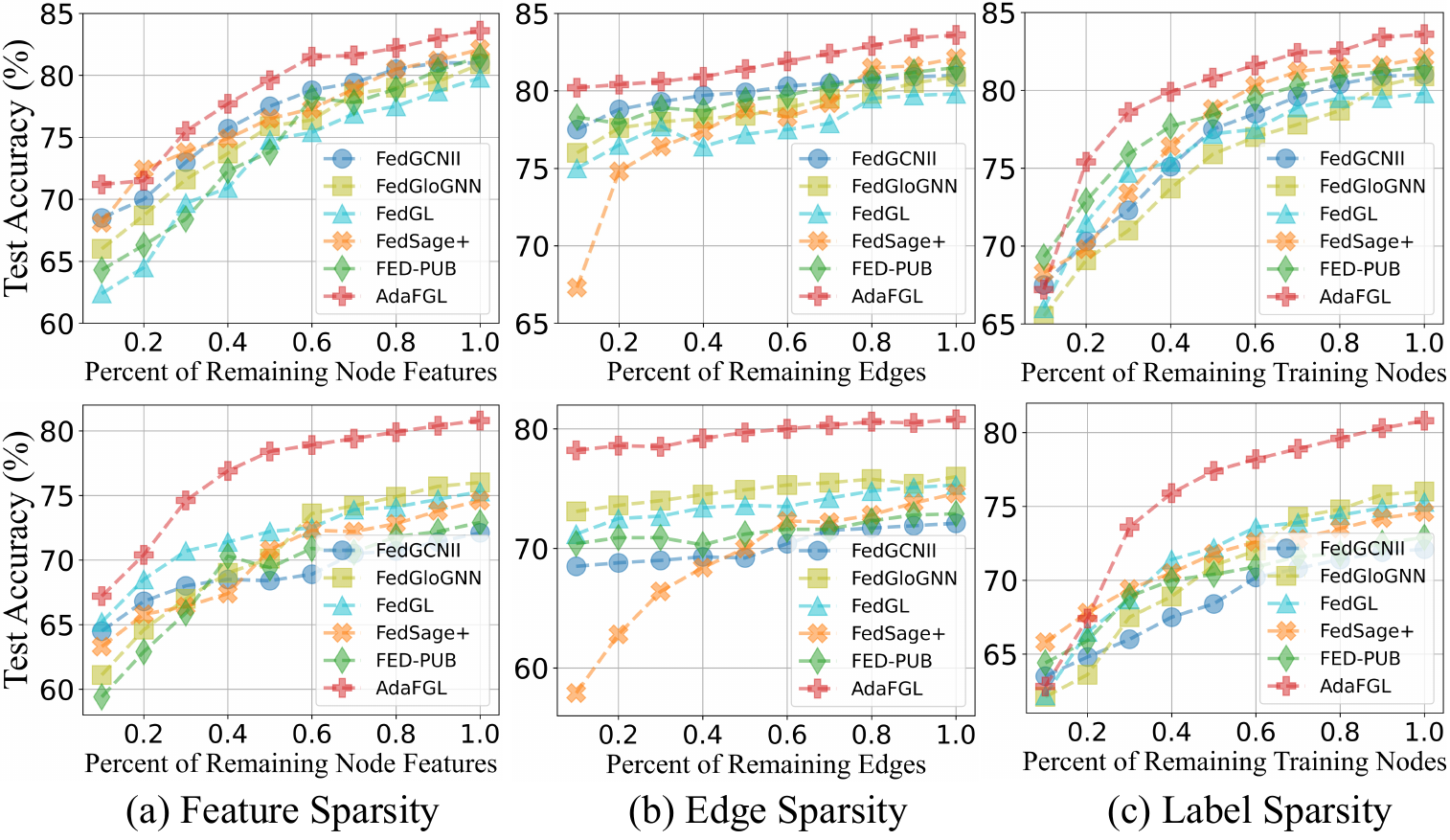}
  \caption{
  {Performance on Computer with community split (upper) and structure Non-iid split (lower) under different levels of sparsity.}}

  \label{fig:exp_sparsity}
\end{figure}

\begin{figure}[t]   
	\centering
    \setlength{\abovecaptionskip}{0.2cm}
    \setlength{\belowcaptionskip}{-0.15cm}
	\includegraphics[width=\linewidth,scale=1.00]{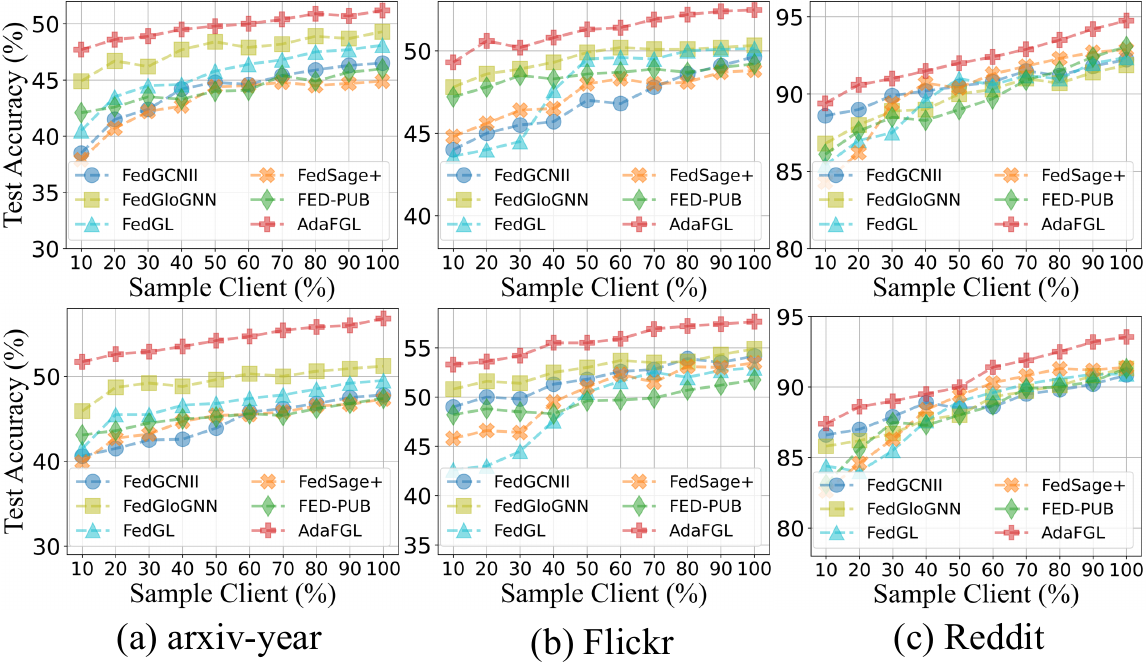}
	\caption{
     {Performance with community split (upper) and structure Non-iid split (lower) under different clients participating.}
    }
    \label{fig: exp_client}
\end{figure}

\subsection{Performance on Sparse Settings}
\noindent \textbf{Sparse Subgraphs.}
    To answer \textbf{Q4}, the experimental results are presented in Fig.~\ref{fig:exp_sparsity}. 
    In the feature sparsity, we assume that the feature representation of unlabeled nodes is partially missing. 
    {
    In this case, Fig.~\ref{fig:exp_sparsity}(a) demonstrates that FedGL and FED-PUB, which rely on self-supervised and model clustering respectively, show sub-optimal performance due to low-quality supervision and underfitting issues.
    Meanwhile, FedGloGNN suffers from the reduced quality of node attribute information due to global embeddings.
    In contrast, FedSage+ and AdaFGL alleviate such issues by generating augmented nodes or utilizing federated knowledge. Such performance can also be applied to edge-sparse and label-sparse scenarios.
    }
    To simulate edge sparsity, we randomly remove edges from subgraphs, providing a more challenging but realistic scenario. 
    For label sparsity, we change the ratio of labeled nodes. 
    Experimental results from Fig.~\ref{fig:exp_sparsity} show that, compared to all existing FGL approaches, our method is more robust to the multi-client subgraphs under different degrees of feature, edge, and label sparsity. 

\noindent \textbf{Sparse Client Participation.}
    In practical FGL scenarios, it is necessary to select a subset of clients to participate in each round to reduce communication costs.
    Thus, we perform 20 client split for arxiv-year, Reddit, and Flickr to present the model convergence performance in Fig.~\ref{fig: exp_client}. 
    {Intuitively, AdaFGL maintains stable performance across different participation ratios due to federated knowledge and local propagation. 
    According to the experimental results, we conclude that the accuracy level of cross-client interaction-based approaches, such as FedGL and FedSage+, significantly drops due to the low participation ratio when subgraphs exhibit high heterogeneity.
    Meanwhile, federated implementations of GNNs like FedGloGNN and FedGCNII are unaffected by fluctuations in the number of participating clients since they do not rely on additional federated optimization.
    In contrast, personalized strategies exhibit robustness, such as AdaFGL and FED-PUB.}

\section{Conclusion}
    {This paper is the first to investigate the topology heterogeneity in federated node classification, aiming to bridge the gap between idealized experimental settings in previous FGL studies and real-world deployment. 
    To provide a unified perspective, we propose a new distributed subgraph simulation strategy called structure Non-iid split, which can be utilized as a benchmark in future FGL studies.}
    Essentially, it generates the topological variance among clients.
    To tackle this challenge, we propose AdaFGL, a user-friendly FGL paradigm that follows a two-step design. 
    It first involves federated collaborative training to obtain a client-shared knowledge extractor, which is then used for personalized training for each client, and the flexibility of propagation modules offers limitless possibilities for the performance boost. 
    A promising future direction is to explore a powerful personalized method by closely integrating it with the globally informed federated model.
    Incorporating the topology and multi-client relationships is also worth exploring.

\section*{Acknowledgment}
 This work was partially supported by 
 (I) the National Key Research and Development Program of China 2021YFB3301301, 
 (II) NSFC Grants U2241211, 62072034, and
 (III) High-performance Computing Platform of Peking University.
 Rong-Hua Li is the corresponding author of this paper.

\newpage
\balance{
\bibliographystyle{IEEEtran}
\bibliography{IEEEexample}
}

\end{document}